\documentclass[lettersize,journal]{IEEEtran}
\usepackage{amsmath,amsfonts}
\usepackage{algorithmic}
\usepackage{algorithm}
\usepackage{array}
\usepackage{subcaption}
\usepackage{textcomp}
\usepackage{stfloats}
\usepackage{url}
\usepackage{verbatim}
\usepackage{cite}
\usepackage{hyperref}
\usepackage{tabularx}
\usepackage{multirow}
\usepackage[switch]{lineno}
\usepackage{booktabs}
\usepackage{newtxtext}
\usepackage{graphicx}
\usepackage{caption}
\usepackage{longtable}
\usepackage{listings}
\usepackage{float}
\usepackage{xcolor}
\usepackage{ragged2e}
\usepackage{orcidlink}
\usepackage{xcolor}
\definecolor{red}{RGB}{153, 21, 78}

\begin{document}

\title{Data Augmentation for Sparse Multidimensional Learning Performance Data Using Generative AI}

\author{Liang Zhang \orcidlink{0009-0002-0017-2569}, Jionghao Lin \orcidlink{0000-0003-3320-3907}, John Sabatini \orcidlink{0000-0002-0292-2039}, Conrad Borchers \orcidlink{0000-0003-3437-8979}, Daniel Weitekamp \orcidlink{0000-0003-0079-8000}, Meng Cao \orcidlink{0000-0002-1286-2885}, John
Hollander \orcidlink{0000-0002-3270-7495}, Xiangen Hu \orcidlink{ 0000-0001-9045-4070}, Arthur C. Graesser \orcidlink{0000-0003-0345-6866}
\thanks{Manuscript submitted on August 4, 2024. First revision on Nov. 23, 2024; Second revision on Dec. 31, 2024. Accepted on Jan.2, 2025. }
\thanks{Liang Zhang, John Sabatini, and Arthur C. Graesser are with the Institute for Intelligent Systems, University of Memphis, Memphis, TN 38152, USA (e-mail: \href{mailto:lzhang13@memphis.edu}{lzhang13@memphis.edu}, \href{mailto:jpsbtini@memphis.edu}{jpsbtini@memphis.edu}, \href{mailto:art.graesser@gmail.com}{art.graesser@gmail.com}).}
\thanks{John Hollander is with Arkansas State University, Jonesboro, AR, 72401, USA (e-mail: \href{mailto:jhollander@astate.edu.}{jhollander@astate.edu}).}
\thanks{Conrad Borchers, Daniel Weitekamp, and Meng Cao are with the Human-Computer Interaction Institute, Carnegie Mellon University, Pittsburgh, PA, 15213, USA (e-mail: \href{mailto:cborcher@andrew.cmu.edu}{cborcher@cs.cmu.edu}, \href{mailto:weitekamp@cmu.edu}{weitekamp@cmu.edu}, \href{mailto:mengc2@andrew.cmu.edu}{mengc2@andrew.cmu.edu}).}

\thanks{Jionghao Lin is with the Faculty of Education, The University of Hong Kong, Hong Kong, PR China, also with the Human-Computer Interaction Institute, Carnegie Mellon University, Pittsburgh, PA, 15213, USA, and also with the Centre for Learning Analytics, Faculty of Information Technology, Monash University, Clayton, VIC 3800, Australia (e-mail: \href{mailto:jionghao@hku.hk}{jionghao@hku.hk}).}

\thanks{Xiangen Hu is with the Department of Applied Social Sciences, Hong Kong Polytechnic University, Hong Kong, PR China (e-mail: \href{mailto:xiangen.hu@polyu.edu.hk}{xiangen.hu@polyu.edu.hk}).}
}

\markboth{Journal of IEEE Transactions on Learning Technologies}%
{Shell \MakeLowercase{\textit{et al.}}: A Sample Article Using IEEEtran.cls for IEEE Journals}


\maketitle
\raggedbottom
\begin{abstract}
Learning performance data, such as correct or incorrect answers and problem-solving attempts in Intelligent Tutoring Systems (ITSs), facilitate the assessment of knowledge mastery and the delivery of effective instructions. However, these data tend to be highly sparse (80\%\(\sim\)90\% missing observations) in most real-world applications. This data sparsity presents challenges to using learner models to effectively predict learners' future performance and explore new hypotheses about learning. This article proposes a systematic framework for augmenting learning performance data to address data sparsity. First, learning performance data can be represented as a 3-Dimensional (3D) tensor with dimensions corresponding to learners, questions, and attempts, effectively capturing longitudinal knowledge states during learning. Second, a tensor factorization method is used to impute missing values in sparse tensors of collected learner data, thereby grounding the imputation on knowledge tracing tasks that predict missing performance values based on real observations. Third, data augmentation using Generative Artificial Intelligence (GenAI) models, including Generative Adversarial Network, specifically Vanilla Generative Adversarial Networks (GAN), and Generative Pretrained Transformers (GPT, specifically GPT-4o), generate data tailored to individual clusters of learning performance. We tested this systemic framework on adult literacy datasets from AutoTutor lessons developed for Adult Reading Comprehension (ARC). We found that: (1) tensor factorization outperformed baseline knowledge tracing techniques in tracing and predicting learning performance, demonstrating higher fidelity in data imputation, and 2) the Vanilla GAN-based augmentation demonstrated greater overall stability across varying sample sizes, whereas GPT-4o based augmentation exhibited higher variability, with occasional cases showing closer fidelity to the original data distribution. This framework facilitates the effective augmentation of learning performance data, enabling controlled, cost-effective approach for the evaluation and optimization of ITS instructional designs in both online and offline environments prior to deployment, and supporting advanced educational data mining and learning analytics.  
\end{abstract}


\section{Introduction}\IEEEPARstart{T}{he} integration of AI-based educational technologies into e-learning platforms, combined with advanced pedagogical strategies, represents a landmark advance in modern education \cite{dominguez2024iot}. This integration has profoundly transformed learning and teaching strategies, making educational systems more adaptive for personalized learning, flexible for remote access, scalable for resource distribution, and effective in improving educational outcomes \cite{tabuenca2024greening}. An exemplary prototype of this advancement is Intelligent Tutoring Systems (ITSs), which provide personalized and adaptive instructions through hints, prompts, and other adaptive feedback to improve learner performance \cite{graesser2018intelligent}. The effectiveness of ITSs has been demonstrated with considerable success in diverse areas such as science, technology, engineering, and mathematics, as well as in fields like reading and language learning \cite{graesser2018intelligent}. Learning performance data, which include records of correct or incorrect answers and problem-solving attempts from learners using the ITS, are crucial for evaluating knowledge mastery, self-regulation, and other characteristics that support adaptive instructions \cite{graesser2004autotutor,koedinger2006cognitive}. For instance, learners' correct and incorrect responses can be leveraged within the learner model component of ITSs to facilitate learner modeling \cite{pavlik2022using,eglington2023optimize}. This process involves tracking historical performance and predicting future outcomes based on knowledge components and concepts, a methodology referred to as Knowledge Tracing (KT) \cite{koedinger2012knowledge}. The ITS can tailor adaptive instructions, such as dialogue-based feedback, based on predictions and assessments, offering targeted support with specific hints and prompts, especially when learners struggle or experience wheel-spinning \cite{graesser2018intelligent}. 

\begin{figure*}[h!t]
\centering 
\includegraphics[width=6.1in]{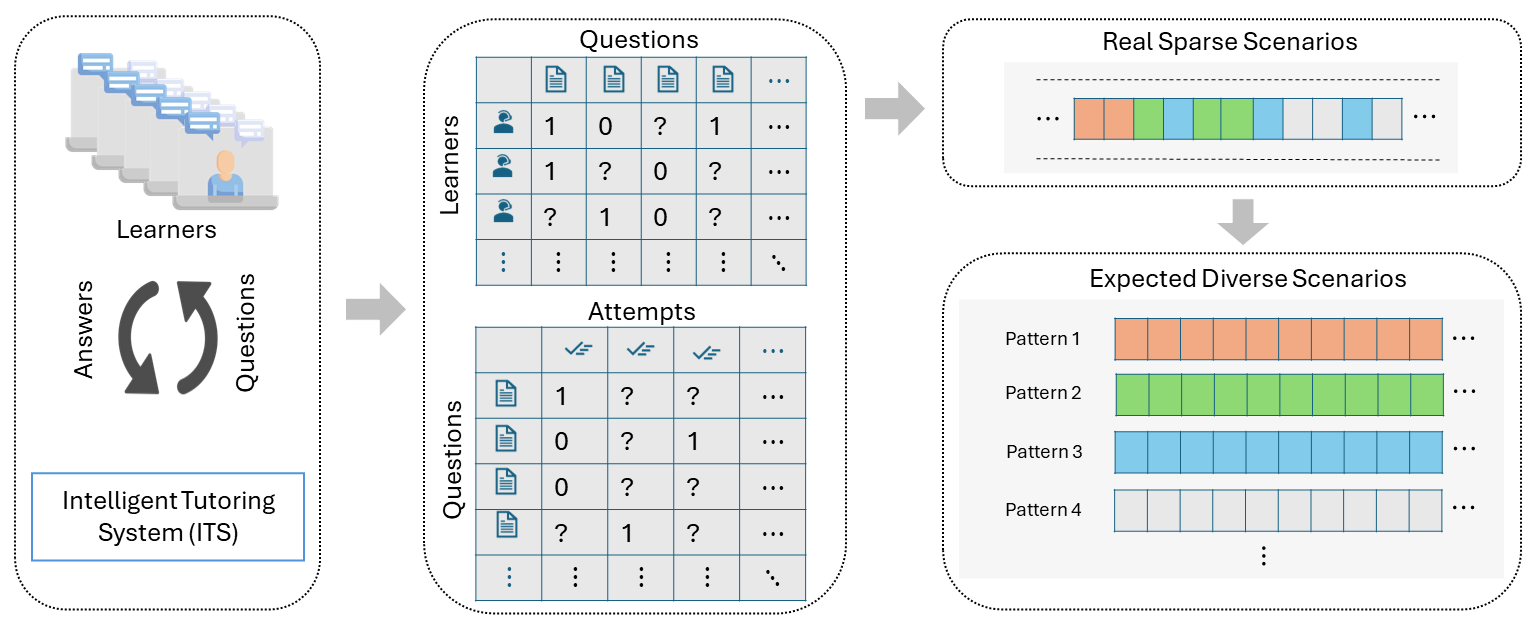}
\caption{Data sparsity issues in learning performance data in Intelligent Tutoring Systems.} 
\label{fig:data_insuf}
\end{figure*}

In real-world educational practices, learning performance data often suffers from data sparsity issues. Fig. \ref{fig:data_insuf} shows how learners interact with an ITS, where their answers and attempts at questions are recorded. The data matrix reveals many missing entries, indicated by the symbol ``\(?\)'', highlighting the sparsity of the data. The real sparse scenarios demonstrate inconsistent and incomplete data patterns, as shown by the different colors within a single bar, representing various biased and unevenly distributed patterns. In contrast, a well-structured and comprehensive dataset is expected to encompass diverse scenarios, with each pattern maintaining relative quantities. This ensures a balanced and thorough representation of the data, facilitating more robust analysis and modeling. Data sparsity is typically divided into two categories: (1) unobserved data resulting from unpresented items (i.e., questions, problems) and limited learner responses or attempts \cite{pandey2019self,lee2022contrastive}, and (2) insufficient learning performance patterns in the available empirical data \cite{wang2023graphca,zhang20243dg}. Insufficient learning performance patterns specifically refer to the lack of quantity in the observed data required to model learning processes effectively in this study. These patterns are often defined by key learner-specific factors, such as initial learning ability (e.g., a learner’s baseline knowledge or proficiency before engaging with new content) and learning rate (e.g., the rate at which a learner improves accuracy or mastery through repeated attempts) \cite{zhang2023exploring,zhang20243dg}. These characteristics not only help distinguish individual or group learning trajectories but also reveal underlying trends, such as whether high-performing learners exhibit slower incremental gains or if low-performing learners make rapid improvements due to starting from a lower baseline. By capturing and analyzing these patterns, learner modeling can identify clusters of learners with similar performance characteristics, thereby enabling more precise and adaptive instructional strategies. Various reasons contribute to these data sparsity issues, including participant dropout from learning tasks \cite{psathas2023predictive}, learner disengagement due to off-task behavior \cite{baker2007modeling} or gaming system \cite{baker2004detecting}, random data loss from design and operation errors \cite{saarela2017automatic}, and potential bias in the experimental groups \cite{greer2016evaluation}. These challenges are typical in real-world settings. 

Data sparsity issues adversely affect the accurate modeling of learning processes and assessment of learners’ knowledge states, which in turn may compromise effective instructional support to learners, particularly those at risk \cite{pandey2019self,wang2023graphca}. Specifically, data sparsity could lead to biased or over-fitted KT models, resulting in potentially misleading predictions of learners’ future performance. Moreover, the insufficiency of learning performance data across a limited range of diverse patterns restricts the potential to comprehensively test and fine-tune ITSs, particularly in cases where certain pre-designed instructional conditions have not been thoroughly explored \cite{baudin2017openturns}. Therefore, addressing data sparsity is crucial for avoiding biased modeling, enabling more comprehensive evaluations of the learning process, and improving effective instructional support in ITSs. 

Addressing data sparsity within ITSs represents a practical yet challenging research area. Informed by advancements in machine learning field \cite{shorten2019survey,emmanuel2021survey}, tackling this issue typically involves two interconnected and sequential strategies: \textit{data imputation} and \textit{data augmentation}. Data imputation focuses on filling the gaps in missing data, thereby creating a more complete dataset, which is crucial for accurate analysis and decision-making \cite{thai2012factorization}. On the other hand, data augmentation not only enriches datasets where learning performance patterns are rarely presented but also enhances the robustness of analysis, modeling, and testing in ITSs, helping to mitigate the effects of sparsity on learner performance prediction \cite{wang2023graphca}. Despite the critical need, systematic efforts to address these data sparsity issues in the ITSs remain limited, with few studies focusing on the comprehensive management of sparse data \cite{thai2012factorization,pandey2019self}.

Effective handling of missing data remains a challenge in educational data analysis. Even though general data imputation methods (e.g., indicator or mean imputation \cite{donders2006gentle}, regression imputation \cite{zhang2016missing}, and multiple imputation \cite{rubin1978multiple}) have proven effective in the literature, they offer a cost-effective solution that avoids labor-intensive experiments by leveraging observed data to estimate unobserved data. These methods capitalize on underlying patterns and characteristics \cite{rubin1977assignment}, but they address the missing data in a straightforward manner that fails to capture the full complexity of the data structure. For example, indicator or mean imputation replaces missing values with a specified indicator or the mean of observed values. However, this method may introduce bias by oversimplifying the complexities of missing data \cite{donders2006gentle}. Regression imputation predicts missing values based on other observed data using regression models. Despite its utility, it often fails to capture the full spectrum of the underlying data structure \cite{zhang2016missing}. Multiple imputation generates multiple datasets by imputing missing values with a range of plausible values and then averaging the results \cite{rubin1978multiple}. Yet, it may not adequately address complex, high-dimensional correlations. Furthermore, data sparsity could influence model prediction of learner performance. When using some KT models in ITSs to predict learning outcomes, such as Performance Factor Analysis (PFA) and Bayesian Knowledge Tracing (BKT), use logistic regression and Bayesian networks, respectively, to predict learning outcomes \cite{pavlik2009performance,yudelson2013individualized}. However, these models themselves are vulnerable to data sparsity and cannot account for the sequence effect of learning events, though partially addressing data sparsity issues \cite{pandey2019self,wang2023graphca}.

Therefore, to effectively address data sparsity and fully capture the data's complexity, and improve the prediction of learner performance, more effective computational approaches that offer deeper insights are needed. In ITS research, tensor factorization, originating from recommendation techniques, has been used to recommend missing performance data based on existing learner records \cite{thai2012factorization}. This method leverages multidimensional relationships to enhance predictions and knowledge representation by integrating three dimensions: learners, items (e.g., questions or learning materials), and temporal factors (e.g., time or attempts) \cite{thai2012factorization, sahebi2016tensor}. Another source of inspiration stems from the advancements of GenAI models, which are capable of generating new data based on patterns during training and have revolutionized data augmentation methodologies by being more flexible and powerful \cite{jovanovic2022generative}. One prototype is Generative Adversarial Networks (GANs), which can proficiently learn from existing data distributions and generate varied samples that extend beyond the original dataset \cite{goodfellow2020generative}. Since its development, the base architecture of GAN, known as Vanilla GAN \cite{goodfellow2014generative}, has inspired numerous variants, such as Deep Convolutional GAN (DCGAN), Conditional GAN (cGAN), and Wasserstein GAN (WGAN). These variants have been widely embraced for their ability to improve stability, handle specific types of data, and generate more realistic samples, making them highly adaptable to various application domains. Another is the Generative Pre-trained Transformer (GPT), which possesses reasoning abilities for learning data distributions and can use both computational and heuristic models to help sample larger datasets that align with the original data distribution \cite{vaswani2017attention}. The evolution of GPT models, from the initial versions such as GPT-2 and GPT-3 to more advanced iterations like GPT-4o and GPT o-1, has brought increasingly sophisticated capabilities, including enhanced contextual understanding and greater fidelity in data generation. These advancements highlight the potential of GPT models to revolutionize data augmentation in domains requiring highly accurate and scalable solutions. Successful applications of GenAI models for data augmentation in ITSs include enriching mathematics student learning datasets with multiple-choice questions \cite{morales2022simultaneous}, generating extensive student behavioral data via GANs, from MOOC learning platforms \cite{li2023sim}, and augmenting sparse learning performance data in reading comprehension \cite{zhang20243dg}. Additionally, leveraging the GPT model has shown significant improvements in selecting appropriate machine learning models and fine-tuning them for predicting learning performance \cite{zhang2024predicting}. Building on the success of GenAI models in ITSs, these models could become highly effective tools for data augmentation, tailored to specific learning needs. 

The present study develops a systematic augmentation framework that integrates multidimensional modeling using tensor factorization for data imputation and GenAI models for data augmentation. This framework generates extensive learning performance data tailored to specific learning performance patterns, enhancing both data imputation and data augmentation capabilities based on model-derived patterns from real-world experimental datasets. The framework will be tested with learning performance data from an example ITS for adult literacy, using AutoTutor lessons for ARC \cite{graesser2016reading}. Our investigation is guided by the overarching \textbf{R}esearch \textbf{Q}uestion: “\textit{How can we develop a data augmentation framework to enhance sparse learning performance data and improve the scalability of learning performance data in the AutoTutor ARC?}” This is explored through the following sub-questions: 

\begin{itemize}
    \item \textbf{RQ1}: How does tensor factorization perform in imputing learner performance data in ARC, particularly in comparison to baseline methods such as PFA, BKT, and Sparse Factor Analysis Lite (SPARFA-Lite)? 
    \item \textbf{RQ2}: How can GenAI models, including GAN specifically Vanilla GAN) and GPT specifically GPT-4o), be effectively and reliably utilized for data augmentation to tailor individual performance patterns? 
\end{itemize}

This study clarifies the selection of two fundamental types of GenAI models as the basis for initializing data augmentation for sparse learning performance data. By employing both GAN-based and GPT-based approaches, we aim to explore how these models can be utilized for data augmentation in learning engineering. Furthermore, we conduct a comparative analysis of these two types of GenAI models to address the challenges posed by data sparsity.

This proposed systematic augmentation framework has the potential to emerge as a powerful tool for enhancing data richness and reliability. Firstly, it can perform data imputation for sparse learning performance data, thereby enriching the data representation for more comprehensive learner modeling. Secondly, the generative models can increase the diversity of individualized learner performance data, which is essential for enriching learner modeling and ITS training. Practically, this study can lead to improved adaptation and responsiveness within ITS environments by providing a broader dataset to fine-tune instructional feedback and interventions. Enhanced data reliability and richness also enable ITSs to make more accurate predictions of individual learner needs, ultimately contributing to personalized learning paths, optimized feedback loops, and better overall learning outcomes. By augmenting a wider range of learner behaviors, the framework supports the continuous improvement of ITS instructional functions, ensuring these ITSs remain adaptable and effective across diverse learner profiles. Our code and some results can be found at the following GitHub link:\href{ https://github.com/LiangZhang2017/3DGAI}{https://github.com/LiangZhang2017/3DGAI}.

\section{Related Work}

\subsection{Intelligent Tutoring Systems for Adult Reading Comprehension}

Adult Reading Comprehension (ARC), essential for academic success and lifelong learning, can be enhanced through AI-based technologies that provide adaptive and personalized solutions \cite{graesser2016reading}. These AI-based technologies leverage educational data mining and learning analytics to assess student performance (e.g., learner modeling), and to provide personalized instruction tailored to individual learners (e.g., tutoring feedback, hints, nudges, and tailored recommendations for reading materials and comprehension exercises) \cite{graesser2017assessment,sabatini2023framework}. For instance, the latest advances in natural language processing and machine learning have enabled digital textbooks to use their content and structure as a knowledge base for ``smart'' functionalities like automated knowledge compilation, adaptive navigation and presentation, and targeted content recommendations \cite{brusilovsky2022return}. Another instance is the AutoTutor developed for ARC, which is investigated in the present study \cite{graesser2016reading}. The AutoTutor ARC employs the trialogue design, which includes one human learner and two computer agents (virtual tutor and virtual companion) \cite{graesser2016reading}. This system facilitates interactive learning through a three-party conversation that assesses learners’ responses, provides feedback, and corrects misconceptions based on an Expectation-Misconception Tailored (EMT) mechanism that attempts to cover good answers (called expectations) and correct bad answers (called misconceptions) \cite{graesser2005computerized}. The tutoring session concludes once all lesson expectations are met. Other ITSs, such as DSCoVAR for vocabulary learning \cite{frishkoff2016dynamic}, and ITSS for reading comprehension instruction through structure strategy training \cite{meyer2016intelligent}, are also noteworthy but are not detailed here. 

Recently, ITSs for ARC have increasingly incorporated advanced AI technologies, particularly GenAI. Large Language Models (LLMs) have already been used to develop summary grading models for intelligent textbooks that provide real-time formative feedback and assess comprehension \cite{morris2024formative}. GenAI models, including GAN and GPT, have been used to tackle sparse data challenges in reading comprehension, thereby enhancing personalized educational technology in ITSs \cite{zhang20243dg}. Furthermore, the use of LLMs to generate high-quality, personalized reading materials underscores the potential for future ITS designs to improve reading comprehension skills \cite{xiao2023evaluating,zhang2024spl}. These AI advancements have revolutionized ITS development from their design and creation to learning analytics and modeling in ARC. 

\subsection{Tensor-based Imputation for Sparse Performance Data}
The increasing prevalence of sparse learning performance data from ITSs necessitates robust imputation methods to handle missing information effectively. Tensor-based imputation has emerged as an important technique due to its ability to maintain the multi-dimensional nature of learning data and to preserve intrinsic relationships across dimensions such as learners, questions, time or attempts \cite{thai2012factorization,sahebi2016tensor,zhang2023exploring}.

Pioneering implementations of tensor-based methods include Thai-Nghe et al. (2011), who extended matrix factorization with tensor factorization to incorporate temporal effects, which significantly enhanced the accuracy of learner performance predictions \cite{thai2012factorization,nghe2015factorization}. Similarly, Sahebi et al. (2016) introduced Feedback-Driven Tensor Factorization (FDTF) to integrate sequences of students, quizzes, and attempts within a tensor framework, which again improved knowledge representation and performance predictions \cite{sahebi2016tensor}. Doan and Sahebi (2019) developed the Ranked-Based Tensor Factorization (RBTF) model that uses tensor factorization to accommodate occasional forgetting of concepts and integrate biases related to students, problems, and time, thereby supporting a predominantly positive learning trajectory \cite{doan2019rank}. Over time, various tensor factorization techniques have been developed to enhance the prediction accuracy and impute sparse data in educational settings \cite{zhao2020modeling,wang2021knowledge}. These developments underscore the capability of tensor factorization to not only enhance prediction accuracy but also perform tensor-based imputation, effectively filling in missing values in sparse tensors. Tensor-based  of learning performance data preserves the ``natural representation'' of synchronous and sequential learning events, which is essential for accurately tracing and predicting learner performance and for improving the imputation of missing values \cite{thai2011factorization}. This process also allows for the efficient decomposition of interactions across different dimensions, enabling deeper analysis and further enriching the dataset’s utility \cite{doan2019rank}. The method’s alignment with recommendation systems highlights its utility in identifying performance similarities and dependencies among learning events, making it a cornerstone in educational data mining. 

\subsection{Generative AI for Augmenting Sparse Educational Data}
The advent of GenAI technologies has revolutionized the field of data augmentation, particularly in domains burdened by sparse and imbalanced datasets \cite{antoniou2017data}. GenAI models have provided groundbreaking ways to synthesize realistic and existing data by capturing complex intrinsic patterns, distributions and characteristics, effectively enhancing dataset robustness for training machine learning models \cite{goodfellow2014generative}. At this point, GenAI effectively addresses scalability challenges in both high and low-dimensional simulation data. It can simulate data on a larger scale and generate tailored scenarios, facilitating informed decision-making under uncertainty conditions arising from sparse datasets \cite{aggarwal2021generative}. Many successful applications in other fields have also exemplified the potential for educational data. Mariani et al. utilized GANs to balance imbalanced image classification datasets \cite{mariani2018bagan}, while Frid-Adar et al. increased the size and diversity of medical imaging datasets through synthetic data augmentation \cite{frid2018synthetic}. Huang et al. transformed images for image-translation tasks, including day-to-night and vice versa, using GANs \cite{huang2018auggan}. In educational settings, GANs have synthesized additional data from sparse datasets in open university learning analytics \cite{hemachandran2022artificial}, illustrating their broad potential for improving learning performance data in ITSs. 

The GPT is a state-of-the-art GenAI model known for its exceptional ability in human-like text generation and performing reasoning tasks with unprecedented accuracy \cite{achiam2023gpt}. ChatGPT represents a significant advancement in AI, driving revolutionary shifts in its application within education. It has been instrumental in enhancing instructional feedback \cite{dai2023can,fung2024automatic}, boosting student engagement \cite{tan2023large}, generating questions \cite{hang2024mcqgen}, simulating learning scenarios \cite{markel2023gpteach}, and providing personalized learning experiences \cite{xiao2023evaluating}. In parallel, recent work highlights ChatGPT’s effectiveness in fostering self-regulated learning \cite{ng2024empowering}, GPT o-1’s potential for advancing higher-order thinking in education \cite{latif2024systematic}, and GPT-based methods that underpin dialogue-driven ITSs \cite{zhang2024spl}. Notably, there has been tremendous progress in data augmentations in ITSs. For instance, Liu et al. \cite{liu2022open} employed ChatGPT to enrich open-ended student responses in computer science with knowledge-guided code for short programming tasks. Zhang et al. \cite{zhang20243dg} demonstrated ChatGPT's capability in learning the distribution of learning performance data tailored to individual patterns and selecting appropriate machine learning models for data augmentation in reading comprehension. Further studies have confirmed that ChatGPT can predict learning performance by encoding datasets, selecting and fine-tuning models, and decoding outputs as estimated probability-based learning performance \cite{zhang2024predicting}. 

All these related works demonstrate the potential of GenAI models for data augmentation in ITSs. By generating synthetic yet realistic examples, these models can significantly expand the depth and breadth of learning performance data, making it more comprehensive and representative. This enables a more accurate assessment of learners' progress and more robust training of AI systems that can adapt to various educational challenges and learner profiles. 

\section{Dataset}

\begin{table*}[t!]
\scriptsize
\caption{Summary of the AutoTutor ARC Lessons Dataset.}\label{tbl1}
\centering
\begin{tabularx}{\textwidth}{@{} p{2cm} p{3cm} >{\centering\arraybackslash}p{2cm} >{\centering\arraybackslash}p{2cm} >{\centering\arraybackslash}p{2cm} >{\centering\arraybackslash}p{2cm} >{\centering\arraybackslash}p{2cm} @{}}
\toprule 
\textbf{Lesson} & \textbf{Topic} & \textbf{Difficulty Level} & \textbf{\#Transactions} & \textbf{\#Learners} & \textbf{\#Questions} & \textbf{\#Max. Attempt} \\ 
\midrule
\multirow{3}{*} {Lesson 1} & \multirow{3}{*}{Persuasive Text} & Medium & 1160 & 107 & 10 & 9 \\ 
 &  & Easy & --- & --- & --- & --- \\ 
 &  & Hard & 854 & 96 & 11 & 8 \\
\midrule
\multirow{3}{*}{Lesson 2} & \multirow{3}{*}{Cause and Effect} & Medium & 1527 & 118 & 9 & 9 \\ 
 &  & Easy & 450 & 48 & 10 & 5 \\
 &  & Hard & 838 & 64 & 10 & 9 \\
\midrule
\multirow{3}{*}{Lesson 3} & \multirow{3}{*}{Problems and Solutions} & Medium & 1444 & 140 & 11 & 5 \\ 
 &  & Easy & --- & --- & --- & --- \\
 &  & Hard & 891 & 124 & 8 & 5 \\ 
\midrule
\multirow{3}{*}{Lesson 4} & \multirow{3}{*}{Inferences from Texts} & Medium & 442 & 46 & 10 & 7 \\ 
 &  & Easy & 102 & 16 & 9 & 2 \\
 &  & Hard & 235 & 25 & 10 & 7 \\
\bottomrule
\end{tabularx}
\end{table*}

Four lesson datasets\footnotemark[1]\footnotetext[1]{AutoTutor Moodel Website: \href{https://sites.autotutor.org/}{https://sites.autotutor.org/}; Adult Literacy and Adult Education Website: \href{https://adulted.autotutor.org/}{ https://adulted.autotutor.org/}} were selected from a total of 29 AutoTutor ARC lessons used in an adult literacy intervention program involving 252 participants. The intervention program spanned from January 2015 to December 2016, with each intervention lasting approximately four months. Ethical approval was obtained from the Institutional Review Board (IRB) under the approval number H15257. As illustrated in Table \ref{tbl1}, the four lessons sourced from the ``\textit{Stories and Texts}'' series within the adult reading comprehension program cover topics including ``\textit{Persuasive Texts}'' (Lesson 1), ``\textit{Cause and Effect}'' (Lesson 2), ``\textit{Problems and Solutions}'' (Lesson 3), and ``\textit{Inferences from Texts}'' (Lesson 4). Each lesson includes 8 to 11 multiple-choice questions to test learners' reading comprehension skills. Learners start with medium (M) difficulty level materials, then, depending on their performance, either progress to the hard (H) level or move down to the easy (E) level. The ``Max. Attempt" column in Table \ref{tbl1} defines the baseline setting for the maximum number of attempts across all selected question records. For example, a sample learning record from the topic ``Persuasive Texts'' involve a anonymous learner identified as ``CSU00032'' (marked as Anon.Student.Id). When asked the question ``What is the topic of the article?'' regarding the main idea of a provided passage, the learner selected the option ``Minimum wage'' by clicking the choice button and received a ``CORRECT'' assessment (as evaluated by AutoTutor) on the first attempt. The ``Transactions'' column in each lesson dataset records the total number of observations, serving as the sample size for each lesson. Note that: 1) It is rare for learners to engage with the easy level in Lesson 1 and 3, since the majority can advance to the hard level; and 2) There are instances of dropout among learners during transitions. For further details on the experiments conducted, see the reference paper \cite{hollander2023importance}.

\IEEEpubidadjcol

\section{Methods}
This section describes our data augmentation framework to enhance the learning performance data for AutoTutor ARC as well as the procedural methods used for its construction. 

\subsection{The Systematic Augmentation Framework}

\begin{figure*}[h!t]
\centering
\includegraphics[width=6.8in]{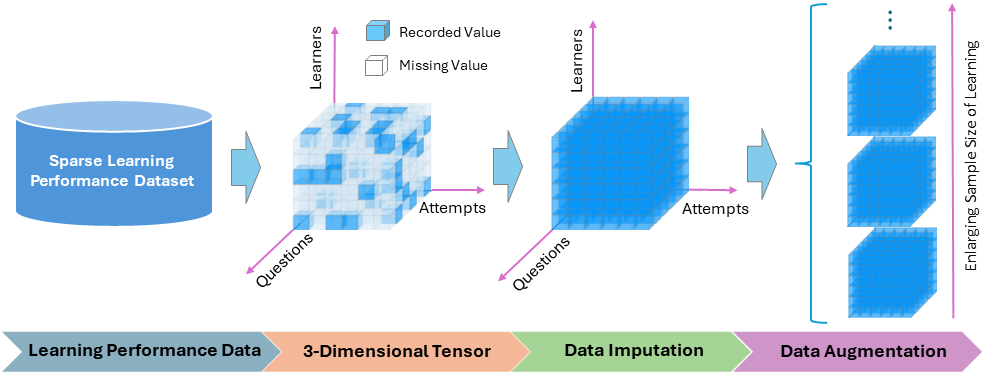}
\caption{The systematic augmentation framework for learning performance in Intelligent Tutoring System.}
\label{fig:framework}
\end{figure*}

The systematic augmentation framework, shown in Fig. \ref{fig:framework} integrates the construction of a 3D tensor to represent learning performance data with subsequent tensor-based imputation and augmentation, effectively enriching these data. Initially, the framework structures learning performance data from real-world learner-ITS interactions into a three dimensional tensor, encompassing learners, questions, and attempts. The entries in this tensor represent learning performance values, quantified as binary values (1 for correct and 0 for incorrect). As depicted in Fig. \ref{fig:framework}, the 3D tensor includes filled cubes that represent recorded learning performance values and transparent cubes that indicate sparse or missing values. Subsequently, tensor-based imputation using Tensor Factorization converts the sparse tensor into a densified form. The densified tensor provides invaluable insights into diverse learning performance patterns, enabling the segmentation of the tensor into sub-tensors based on these distinct patterns (this will be detailed in subsequent sections). GenAI models, such as GAN and GPT, are then used to simulate additional data samples, enriching the original dataset based on specific learning performance patterns. This scalable simulation process ultimately generates a more comprehensive dataset that incorporates both imputed and augmented data.

The workflow framework can be formalized as follows:
\small
\begin{equation} logs_{learning} \rightarrow\boldsymbol{\mathcal{T}}_{sparse} \rightarrow \boldsymbol{\mathcal{T}}_{dense} \rightarrow \boldsymbol{\mathcal{T}}_{augmented} 
\end{equation}
where \(logs_{learning}\) represents the logs of learning performance (including learner, questions, attempts, and actual learning performance values). These logs are structured into sparse tensor \(\boldsymbol{\mathcal{T}}_{sparse}\). This sparse tensor is then processed through tensor-based imputation techniques to yield \(\boldsymbol{\mathcal{T}}_{dense}\), a densified tensor that hat fills in missing or sparse values to create a more complete dataset. The densified tensor is then used to further augment the tensor into \(\boldsymbol{\mathcal{T}}_{augmented}\), scaling datas sample by individualized learning performance patterns. 

\subsection{Construction of 3D Tensor for Learning Performance Data}

Consider the data produced by a population of learners working in an ITS. The data consist of a set of \(U\) learners, represented by \(\{l_1,l_2,l_3,\cdots,l_U\}\), who engages with a sequence of \(N\) questions, denoted by \(\{q_1,q_2,q_3,\cdots,q_N\}\). Each question in this sequence permits up to \(M\) attempts, represented by \(\{t_1,t_2,t_3,\cdots,t_M\}\), allowing individual learners multiple opportunities to respond. The \(\boldsymbol{\mathcal{T}}_{sparse} \in R^{U\times N \times M } \) captures these interactions, where the entry \(\tau_{uij}\) in \(\boldsymbol{\mathcal{T}}_{sparse}\) records the performance of learner \(l_u\) on question \(q_{i}\) at the attempt \(a_j\), where the \(u \in(1,2,\cdots,U)\), \(i\in(1,2,\cdots,N)\), \(j\in(1,2,\cdots,M)\). Within the AutoTutor ARC context, the entry variable \(\boldsymbol{\tau}_{uij}=\{0,1, NaN\}\), where 1 indicates a correct answer, 0 signifies an incorrect answer, and \(NaN\) donates unobserved values. 

The construction of the 3D tensor is guided theoretically by the following assumptions: \textbf{(a)} \textit{Hierarchical Knowledge Representation:} Each question involves distinct knowledge components or concepts, including both procedural and declarative types; these knowledge components may also be shared across different questions \cite{corbett1994knowledge,zhao2020modeling}. \textbf{(b)} \textit{Latent Knowledge Relations:} The specific knowledge embedded within each question is crucial for mastery, and the unique yet interconnected knowledge across questions create a comprehensive network of logic and procedures during the knowledge acquisition process \cite{koedinger2012knowledge,wang2021knowledge}. \textbf{(c)} \textit{Sequence Effects in Performance Interactions:} Learners’ sequential interaction with questions affects their understanding, comprehension, and performance; for example, one question might facilitate comprehension and performance of a subsequent question \cite{conway2001sequential,ramscar2016learning} and there may be recency effects where performance in recent items may be weighted higher on learner mastery \cite{pavlik2021logistic,wang2021knowledge}. \textbf{(d)} \textit{Maximum Attempt Assumption}: Assume a theoretical maximum number of attempts a learner might need, highlighting the importance of evaluating comprehensive learning states through repeated attempts. \textbf{(e)} \textit{Similarity in Learning for Individual Learners}: Assuming a common relevance and utility in the mode of knowledge acquisition among learners, it becomes possible to predict knowledge mastery based on similarities in their individual learning pathways \cite{thai2011factorization}. \textbf{(f)} \textit{Probability-based Prediction}: Predicted learning performance is represented as a probability from 0 to 1, indicating the likelihood of knowledge mastery.

\subsection{Tensor-based Imputation}

We model the sparse tensor \(\boldsymbol{\mathcal{T}}_{sparse}\) through factorization into two lower dimensional components: (1) a factor matrix \(\boldsymbol{\mathcal{U}}\) of size \textbf{\(U\times K\)}, which captures the latent learning-related features of \(U\) learners, such as initial learning abilities and learning rates, where \(K\) is the total number of these features; and 2) a latent tensor \(\boldsymbol{\mathcal{V}}\) of size \(K\times M \times N\), representing learner knowledge in terms of \(K\) latent features across \(M\) attempts on \(N\) questions. The approximated tensor \(\boldsymbol{\mathcal{T}}_{dense}\) is  is computed as follows:
\small
\begin{equation} 
\boldsymbol{\mathcal{T}}_{dense} \approx \boldsymbol{\mathcal{U}} \times \boldsymbol{\mathcal{V}} + b_l + b_a + b_q + \varepsilon
\end{equation}
where \(b_l\), \(b_q\), and \(b_a\) represent the biases from learners, questions, and attempts, respectively, and \(\varepsilon\) denotes a global bias. During tensor factorization, the sigmoid function is applied to normalize the output of estimated performance values, ensuring outputs are bounded between 0 and 1. The model also incorporates a rank-based constraint, which promotes a trend of monotonic knowledge acquisition across successive attempts by learners, while still allowing for potential forgetting or slipping \cite{doan2019rank}. The objective function includes terms for mean squared error to measure the discrepancy between observed values and predictions, and regularization terms for the decomposed components (the learner feature matrix \(\boldsymbol{\mathcal{U}}\) and the latent tensor \(\boldsymbol{\mathcal{V}}\) and various biases) to mitigate overfitting, along with a rank-based constraint. Optimization is carried out using stochastic gradient descent to minimize the objective function, iterating until convergence. Ultimately, the resulting \(\boldsymbol{\mathcal{T}}_{dense}\) functions as a probability-based filled tensor. 

The Tensor Factorization method thus enables us to construct and assess different multidimensional models for learning performance that facilitate data imputation for comprehensive modeling of the learning process. We explored three different baseline models that will ultimately be compared with our proposed model. \textbf{(a)} \textit{Bayesian Knowledge Tracing (BKT):} BKT uses a Hidden Markov Model to dynamically assess and predict a learner's knowledge state (represented as binary states of ``known'' and ``unknown''), with adjustments based on their responses to questions while considering probabilities of learning, guessing and slipping \cite{pardos2010modeling}; here, it also initially integrates both student-specific and skill-specific parameters to effectively account for individual learner variability and the hierarchical nature of skills \cite{pardos2010modeling,yudelson2013individualized}. \textbf{(b)} \textit{Performance Factor Analysis (PFA):} PFA utilizes logistic regression to estimate the probability of the learner's performance on the question, factoring in individual learning ability, skill-related features (e.g., difficulty), and the learner's previous success and failures \cite{pavlik2009performance}. \textbf{(c)} \textit{Sparse Factor Analysis Lite (SPARFA-Lite):} SPARFA-Lite, a streamlined version of Sparse Factor Analysis, uses matrix completion techniques to efficiently analyze graded learner responses and predict performance by determining the optimal number of knowledge components, offering enhanced computational speed over the initial Sparse Factor Analysis \cite{lan2014quantized}. The choice of these baseline models was guided by two primary criteria. First, we prioritized interpretability. Complex deep knowledge tracing models, while potentially more accurate, were excluded due to their lack of transparency, which could hinder clear explanation and analysis of results. Second, although prediction accuracy was an important consideration, it was not our primary focus. This study emphasizes addressing data sparsity issues, aiming to provide actionable insights into effectively handling sparse data. While alternative methods may enhance prediction accuracy, many do so by relying on sparse data without directly addressing the underlying sparsity challenges, which falls outside the scope of this investigation. 

\section{Identification of Learning Performance Patterns by Clustering}

To make data augmentation adaptive to individual learning performance patterns, we first identify these patterns by clustering based on the similarity of each learner's performance across different attempts at specific questions. 

The matrix slice \(\boldsymbol{\Omega}_{q_n}\), extracted from the \(\boldsymbol{\mathcal{T}}_{dense}\), encapsulates the probability-based knowledge states associated with the performance on the \(n^{th}\) question \(q_n\), for all \(U\) learners over \(M\) attempts. Eq.~\ref{eq:clusterModelPara} illustrates our approach to identifying learning performance patterns within \(\boldsymbol{\Omega}_{q_n}\). The matrix can be represented as a sequence of vectors \(\{\boldsymbol{L}_1, \boldsymbol{L}_2, \boldsymbol{L}_3, \cdots, \boldsymbol{L}_n\}\), where each \(\boldsymbol{L}_u\) aggregates the \(u^{th}\) learner's performance across all attempts for the specified question. The performance distribution for each learner vector is assumed to be modeled by function \(\boldsymbol{\mathcal{G}}(\cdot)\), with the associated set of parameter vectors depicted as \(\{\boldsymbol{\vartheta}_1,\boldsymbol{\vartheta}_2,\boldsymbol{\vartheta}_3,\cdots,\boldsymbol{\vartheta}_M\}\). The fluctuations in these model parameters indeed reflect individual differences in learning performance patterns by quantifying the uncertainties within the evolving knowledge states for each learner.
\small
\begin{equation} \label{eq:clusterModelPara}
\footnotesize
    \begin{aligned}
        \boldsymbol{\Omega}_{q_n} &= \left\{ \begin{array}{c} 
            \boldsymbol{L}_1\\
            \boldsymbol{L}_2\\
            \boldsymbol{L}_3\\
            \vdots\\
            \boldsymbol{L}_U
        \end{array}\right\} \Rightarrow \left\{ \begin{array}{c}             \mathcal{G}_1(\boldsymbol{L}_1) \\            \mathcal{G}_2(\boldsymbol{L}_2) \\
 \mathcal{G}_3(\boldsymbol{L}_3) \\
            \vdots \\
            \mathcal{G}_{C}(\boldsymbol{L}_U) 
        \end{array} \right. \Rightarrow \left \{ \begin{array}{c}
            \boldsymbol{\vartheta}_1 \\
            \boldsymbol{\vartheta}_2 \\
            \boldsymbol{\vartheta}_3 \\
            \vdots \\ \boldsymbol{\vartheta_{U}} 
        \end{array}\right. \Rightarrow \left \{ \begin{array}{c}
            Cluster_1 \\
            Cluster_2 \\
            Cluster_3 \\
            \vdots \\
            Cluster_{C} 
        \end{array}\right. 
    \end{aligned}
\end{equation} 

In this study, we employed a power law function for \(\boldsymbol{\mathcal{G}}(\cdot)\) to model the relationship between learners' performance values and their number of attempts, drawing on the learning curve theory proposed by Newell and Rosenbloom \cite{newell1980mechanisms}. This theory links error rates to practice amounts and supports the power law's use \cite{newell1980mechanisms}. The power law learning curve is particularly valued for its robust fit and interpretable parameters, which has been widely recognized in educational and training research \cite{dekeyser2020skill}. In the power-law formula \(Y=aX^{b}\), \(Y\) represents learning performance, quantified as the probability of producing correct answers, while \(X\) denotes the number of attempts to respond to the current question. The parameter \(a\) indicates the learner's initial learning ability or prior knowledge, and \(b\) quantifies the rate at which the learner acquires knowledge through practice. We then utilized K-means++ algorithm \cite{arthur2007k} to cluster the distribution of these parameters (\(a\) and \(b\)) among learners, which assists in identifying distinct individual learning performance patterns. 

\section{Data Augmentation based on Generative Models}
We investigated two generative AI models, GAN and GPT, to generate learner data that manifests particular performance patterns, and thereby enable scalable sampling. 

\textbf{Generative Adversarial Networks.} In this study, we employ the Vanilla Generative Adversarial Network (GAN) \cite{goodfellow2014generative}, which represents the foundational architecture in the GAN family. It consists of two complementary networks: a generator (\textit{\textbf{G}}) and a discriminator (\textit{\textbf{D}}), both implemented using fully connected (dense) layers \cite{goodfellow2014generative,goodfellow2020generative}. The generator is designed to produce simulated data samples from initialized random noise, typically sourced from a Gaussian distribution. Its output is crafted to be compatible with the input from the individualized learning performance pattern as synthetic sample data. The discriminator's role is to determine whether the augmented data samples are real or synthetic by comparing them with actual data samples (original learning performance distribution). Concurrently, the \(\boldsymbol{G}(\cdot)\) is trained to progressively reduce the difference between the distributions of the real and augmented data through iterative tuning. The training costs for both \(\boldsymbol{D}(\cdot)\) and \(\boldsymbol{G}(\cdot)\) are dictated by the objective function \(V({\boldsymbol{G},\boldsymbol{D}})\), defined as follows \cite{goodfellow2014generative}:
\small
\begin{equation}
\footnotesize
\begin{gathered} \label{ValueFunction}
    \underset{\boldsymbol{D}}{max}\ \underset{\boldsymbol{G}}{min}\ V({\boldsymbol{G},\boldsymbol{D}}) =
    E_{simulate}[log\boldsymbol{D}(\boldsymbol{\mathcal{T}}_{real},\boldsymbol{\mathcal{T}}_{simulate})]+ \\ E_{noise}[log(1-D(\boldsymbol{G}(Random \ Noise)))]
\end{gathered}
\end{equation}
where the \(\boldsymbol{\mathcal{T}}_{real}\) represents the real performance sample input to \(\boldsymbol{G}(\cdot)\), and \(\boldsymbol{\mathcal{T}}_{simulate}\) denotes the output \(\boldsymbol{G}(\cdot)\), simulating real data for assessment by \(\boldsymbol{D}(\cdot)\). The term \(E_{simulate}[\cdot]\) calculates the expectation of log-probability that \(\boldsymbol{D}(\cdot)\) correctly identifies whether data is real or simulated, while \(E_{noise}[\cdot]\) measures the expectation of log-probability that \(\boldsymbol{D}(\cdot)\) correctly rejects generated data as fake. In practical implementations, this Vanilla GAN-based augmentation approach takes the matrix slice \(\boldsymbol{\Omega}_{q_n}\), which represents the probability-based learning performance for learners across all attempts for the \(n^{th}\) question \(q_n\), as the real \(\boldsymbol{\mathcal{T}}_{real}\). By training the Vanilla GAN on the learner-based vectors data from \(\boldsymbol{\Omega}_{q_n}\), the learner dimension can be scaled, enabling synthetic augmentation of the existing data. The Vanilla GAN undergoes training over multiple epochs (N=3,000), alternating updates between the generator and discriminator. Each epoch begins with noise input for the generator, which produces synthetic data at a predefined sample size for learners. The discriminator is then trained first on real data, labeled as 1, and then on synthetic data, labeled as 0, with an average loss calculated to assess its accuracy in distinguishing real from synthetic samples. Subsequently, the generator is trained by propagating the gradient from the discriminator back to the generator. Specifically, the generator takes a random noise vector as input and generates synthetic data, which is then fed to the discriminator. The discriminator evaluates this synthetic data and assigns a probability score indicating whether the data is real or fake. The generator’s loss is computed based on how effectively it ``fools'' the discriminator into classifying the synthetic data as real. This is achieved by minimizing the discriminator's output loss for the synthetic data. Using this loss, the generator's parameters are updated via backpropagation using gradient descent. Over successive epochs, the generator learns to produce increasingly realistic data that aligns closely with the statistical patterns of the real learner performance data. Ultimately, the finely tuned generator produces scalable augmented sample data. 

\begin{table*}[ht!]
\caption{Example prompt strategy for augmenting sparse learning performance using GPT-4o.}
\label{tb:example_prompt}
\scriptsize
\centering
\renewcommand{\arraystretch}{1.3}
\begin{tabular}{>{\raggedright}p{9.0cm} | >{\raggedright\arraybackslash}p{7.5cm}}
\hline
\textbf{User Input} & \textbf{GPT-4o Output} \\
\hline

\textbf{User:} 
Hello ChatGPT, your expertise as an AI language model makes you ideal for assisting me in generating synthetic learning performance data for individual learners. The primary task involves a thorough analysis of the existing learner-attempts matrix, where each value indicates the probability of a learner correctly answering a specific question on a particular attempt. We want to scale this data to accommodate more learners. 

\smallskip
\textit{(Provides relevant reading material, question information, and sample JSON.)} 

\smallskip
...
&
\textbf{GPT-4o:} 
To generate synthetic learning performance data for individual learners, we will:
\begin{itemize}
    \item Analyze the current distribution of probabilities in the learner-attempts matrix.
    \item Define parameters to mimic real-world performance patterns (e.g., power-law distributions).
    \item Generate the new, larger synthetic dataset that retains the original data’s statistical properties.

...

\end{itemize}
\\
\hline
\textbf{User:} \{You can ask some questions to test if GPT-4o truly understands the information provided.\}
& 
\textbf{GPT-4o:} \{Present the results based on your testing query.\}
\\
\hline
\textbf{User:}
The current learners-attempts matrix, partially shown as:
\[
\begin{bmatrix}
0.99759290 & 0.99975013 & 0.99962294 & \dots \\
0.99932609 & 0.99996001 & 0.99990710 & \dots \\
0.99705196 & 0.99976564 & 0.99956506 & \dots \\
\vdots & \vdots & \vdots & \ddots
\end{bmatrix}
\]
corresponds to the question “What is the topic of the article?” (choices: [“Sports”, “Exercising”, “Skydiving”], correct answer: “Exercising”). Each matrix cell indicates the probability (ranging from 0 to 1) of a learner correctly answering on each attempt. These probabilities follow a power-law function with increasing attempts. In addition, the matrix uses the k-means++ algorithm to categorize learners by their fitted power-law parameters, although all learners fell into one cluster due to similar performance patterns.

Please learn from this matrix and generate a new, augmented set of 1000 learners (i.e., 1000 rows) with 9 attempts (i.e., 9 columns). In doing so:
1) Use your computational capability to simulate the augmented matrix.
2) Present the final result as a complete matrix in NumPy array format.
3) Avoid including instructions, explanations, or extra text in the output.
4) Provide the entire matrix as a downloadable CSV file. ...
&
\textbf{GPT-4o:}
The augmented learners-attempts matrix for a population size of 1000 learners has been generated and saved as a CSV file. ...

\\
\hline

\textbf{User:} 
Can you provide details about the model and setup you used for generating the synthetic data? ...
&
\textbf{GPT-4o:}
For generating the synthetic learners-attempts matrix, I employed the following model and setup:
\begin{itemize} 
    \item[\(\bullet\)] \textit{Model details and parameters}
    \item[\(\bullet\)] \textit{Software libraries and environment}
    \item[\(\bullet\)] \textit{Key assumptions or constraints}
\end{itemize}
Upon executing the code, the synthetic learners-attempts matrix was successfully generated. ...

\\
\hline

\textbf{User:}
How can I fine-tune the model to achieve more realistic augmented data?
...
&
\textbf{GPT-4o:}
To fine-tune the sampling model for data augmentation and increase the realism and variability of the synthetic data, consider:
\begin{itemize}
    \item[\(\bullet\)] Introducing learner-level parameter variations 
    \item[\(\bullet\)] Adjusting sampling distributions to match observed performance trends
    \item[\(\bullet\)] Incorporating noise or slight randomization to replicate real-world fluctuations
\end{itemize}

These enhancements ensure that the generated data more closely reflects authentic learner behavior. 
...
\\
\hline

\textbf{User:}
...
&
\textbf{GPT-4o:} 
...
\\
\hline

\end{tabular}
\end{table*}

\textbf{Generative Pre-trained Transformer.} This study highlights the use of GPT-4o for data augmentation, which involves capturing the actual data distribution and generating simulated data samples through a strategic, prompt-based process. This process, also detailed in Zhang et al. \cite{zhang20243dg,zhang2024predicting}, involves three main procedures: encoding, which converts numerical learning performance values into contextual prompts; an LLM component that involves prompting an augmentation method that capitalizes on GPT’s reasoning and understanding capabilities; and decoding, which entails generating the simulated data along with interpretations. In this study, we used GPT-4o, and all further references to GPT indicate this version. As for the aforementioned matrix slice \(\boldsymbol{\Omega}_{q_n}\) that represents the learners and questions for learning performance, it can be formalized as \(GPT(\boldsymbol{\Omega}_{q_n})\), which scales the size of the learner's data. Specifically, inputs such as individualized learning performance values are contextualized with detailed information about questions, answers, and attempts, including descriptions of their format and content. Subsequently, a simulation request prompts GPT-4o to seamlessly integrate this numerical and textual data, driving the execution of a simulation. During this process, sampling mathematical models can be searched, selected, and fine-tuned by GPT-4o to enhance the data augmentation process. The prompts are iteratively refined to yield results that align with our specifications. Additionally, the Chain-of-Thought (CoT) prompting technique \cite{wei2022chain} is incorporated, which uses guiding sentences such as \textit{``Can you explain your understanding of this data?''}, \textit{``Can you suggest potential machine learning methods for augmenting the data, particularly concerning sample size of ... ?''}, and \textit{``Could you provide the results and settings of the model?''}. These prompts are designed to facilitate a more structured simulation process by encouraging GPT-4o to think through each step systematically. Table \ref{tb:example_prompt} shows a sample conversation that demonstrates the step-by-step augmentation process for learning performance data. By embedding numerical data in carefully crafted prompts and drawing upon GPT-4o’s CoT, we iteratively generate an expanded learners-attempts matrix that retains critical statistical features. This approach underscores the flexibility and adaptability of GPT-based methods in capturing subtle learning patterns. Furthermore, prompt refinements and domain-specific guidance allow GPT-4o to introduce appropriate variability while preserving the core behaviors of the original dataset. As a result, the method offers a robust way to scale data for broader analyses without compromising fidelity.  

\section{Experimental Setup and Evaluation}
The experimental setup is designed and optimized according to the framework's workflow, allowing for thorough evaluation through detailed procedures.
\begin{itemize}
    \item \textbf{(a)} \textit{Measurement of Sparsity Levels:} The evaluation of sparsity levels is conducted by calculating the missing rate, defined as the percentage of missing values relative to the total entries in the tensor of learning performance data \cite{thai2012predicting}. 
    \item \textbf{(b)} \textit{Latent Features from Tensor Factorization:} The number of latent features can range from 1 to 20, and the optimal setting is identified through a grid search method that involves iterative optimization of Tensor Factorization in data modeling. 
    \item \textbf{(c)} \textit{Question-level Knowledge Component Configurations:} The BKT and PFA models utilize two distinct assumptions for setting KCs: ``Single KC'', where all questions are attributed to one common knowledge component, and ``Unique KC'', which assumes a one-to-one correspondence between questions and knowledge components. 
    \item \textbf{(d)} \textit{Cross-Validation for Running Data Imputation Models:} For Tensor Factorization and other baseline models in the data imputation stage, training and testing consistently follow a 5-fold cross-validation strategy to ensure reliable results, with all models evaluated by averaging these outcomes or through five independent runs for enhanced robustness. 
    \item \textbf{(e)} \textit{Evaluation of Predictive Accuracy for Data Imputation:} As referenced in other peer research \cite{doan2019rank,wang2021knowledge,yudelson2013individualized,xiong2010temporal}, both Mean Absolute Error (\(MAE\)) and Root Mean Square Error (\(RMSE\)) are employed to quantify the performance of the models. 
    \item \textbf{(f)} \textit{Significance of Model Performance by ANOVA Analysis:} ANOVA analysis is employed to assess the statistical significance of model performance, using \textit{MAE }and \textit{RMSE} to critically evaluate and compare the effectiveness of tensor factorization against other models. 
    \item \textbf{(g)} \textit{Scaling Sample Size:} Varying sample sizes for data augmentation are compared between Vanilla GAN and GPT-4o models to examine their effects on variance and bias in parameter estimates fitted to augmented learner performance datasets. Sample sizes increase in increments of 1,000, ranging from 1,000 to 20,000.   
    \item \textbf{(h)} \textit{Divergence Measurement:} The Earth Mover's Distance (\textit{EMD}) metric \cite{rubner2000earth,pele2009fast} is employed to quantify divergence by measuring the distance between parameters of the power-law functions fitted to augmented and original data distributions separately, and to assess the reliability of data augmentation across varying sample sizes. Before computing \textit{EMD}, the parameters are normalized within each sample size group to ensure comparability, as normalization scales the distributions to sum to 1, removing biases due to differences in magnitude or sample size. In this study, we use this normalization method to scale values while preserving their relative proportions. Instead of considering other methods like Min-Max normalization, which may distort the distribution’s shape, or Z-Score normalization, which is unsuitable for probability-based comparisons, we chose this approach to ensure the data remains a valid probability distribution. \textit{EMD} measures the minimum cost of transforming one distribution into another, reflecting the amount of work needed. Smaller \textit{EMD} values indicate closer similarity between distributions, while larger values indicate greater divergence. This \textit{EMD} metric effectively captures how changes in simulation sample size influence divergence from the original data. This procedure is formalized as follows. Firstly, normalize the original distribution \(
   \hat{o}_i = \frac{o_i}{\sum_{j=1}^{n} o_j}, \quad \forall i \in \{1, \ldots, n\}
   \), and the augment distribution \(
   \hat{s}_i = \frac{s_i}{\sum_{j=1}^{m} s_j}, \quad \forall i \in \{1, \ldots, m\}
   \). Next, compute the EMD using the following equations: 
   \small
   \begin{equation}
   C_{\hat{O}}(i) = \sum_{j=1}^{i} \hat{o}_j, \quad C_{\hat{S}}(i) = \sum_{j=1}^{i} \hat{s}_j
   \end{equation}
   \small
   \begin{equation}
   W_1(O, S) = \sum_{i=1}^{n} \left| C_{\hat{O}}(i) - C_{\hat{S}}(i) \right| \cdot d_i
   \end{equation} 
   where the \( d_i \) represents the distance between successive points (often assumed to be 1 if indices correspond to bins or parameters), and \( C_{\hat{O}}(i) \) and \( C_{\hat{S}}(i) \) are the cumulative sums of the normalized vectors up to index \( i \), ensuring an accurate calculation of the \textit{EMD}.
   \item \textbf{(h)} \textit{Evaluations of Parameters Distributions Characteristics}: To comprehensively evaluate the augmented datasets, we employed supplementary measures, including visualization techniques (violin plots) and quantitative metrics like the Interquartile Range (IQR) \cite{wan2014estimating} and Bimodality Coefficient (BC) \cite{pfister2013good}, alongside the EMD for divergence measurement. Violin plots were utilized to illustrate the distribution shapes and spread of parameters \(a\) and \(b\), providing detailed insights into central tendencies and variability in both Vanilla GAN and GPT-4o augmented data relative to the original dataset. The IQR, defined as the range between the 25th percentile (first quartile) and the 75th percentile (third quartile) of a distribution, was calculated as a robust, nonparametric measure of spread. Unlike standard deviation, the IQR effectively captures variability while reducing the influence of outliers, making it particularly suited for comparing distributions with differing levels of skewness or kurtosis. The BC was calculated based on the skewness and kurtosis of the parameter distributions to quantitatively assess their modality. Additionally, the BC was employed to quantitatively assess the modality of the distributions, The formula for BC is: 
   \begin{equation}
       BC = \frac{g^2 + 1}{k + \frac{3(n-1)^2}{(n-2)(n-3)}}
   \end{equation}
   where \(g\) represents the skewness, \(k\) is the kurtosis, and \(n\) is the sample size. The threshold for bimodality was set at \(BC=0.555\), with values above this threshold generally indicating bimodality, while values below suggest unimodality. This coefficient integrates asymmetry and peakedness to provide a comprehensive measure of the distribution’s modality. By integrating these supplementary measures, we performed a rigorous analysis of the structural fidelity and variability of the augmented data, enabling a comprehensive assessment of the augmentation models' performance. 
\end{itemize}

To achieve optimal models' performance, it's essential to fine-tune the parameters of each model during the optimization process. For the BKT, the four key parameters (\(P(L_0)\), \(P(S)\), \(P(G)\) and \(P(T)\)) start with initial values within a range of 0.05 to 0.95. The optimization process tailors these parameters to each specific KC, demonstrating their adaptability to the unique characteristics of each KC. The PFA model using the generalized linear mixed model with individual learners as random effects for each KC (under ``Unique KC'' mode) emphasizes the consideration of individual differences among different learners and their related skills for acquiring KCs in the learning process. SPARFA-Lite derives its original matrix by averaging performance across multiple attempts, a method that simplifies the data preparation step. The Tensor Factorization, which is applied to various AutoTutor ARC lessons, necessitates a different set of tuning parameters, including \(\lambda\), \(\lambda_1\), \(\lambda_2\), \(\eta\), \(K\) (as detailed in Table~\ref{tb:sparse}), and learning rate \(lr\). In our experiments, \(\lambda\) was adjusted from \(10^{-1}\) to \(10^{-6}\), \(\lambda_1\) and \(\lambda_2\) ranged from \(10^{-4}\) to \(10^{-9}\), and \(lr\) varied from 0.5 to \(10^{-2}\). 

\section{Results}
\subsection{Sparsity Measurement and Latent Features Obtained by Tensor Factorization}

\begin{table*}[ht!]
\caption{Comparative analysis of model prediction accuracy in AutoTutor ARC lesson dataset using \(RMSE\). Bold values represent the lowest values.}\label{tbl_rmse}
\centering
\begin{tabularx}{\textwidth}{@{}p{1.9cm} >{\centering\arraybackslash}p{2.2cm} >{\centering\arraybackslash}p{2.3cm} >{\centering\arraybackslash}p{2.2cm} >{\centering\arraybackslash}p{2.2cm} >{\centering\arraybackslash}p{2cm} >{\centering\arraybackslash}p{2.6cm}@{}}
\toprule
\textbf{Dataset} & \textbf{BKT (Single KC)} & \textbf{BKT (Unique KC)} & \textbf{PFA (Single KC)} & \textbf{PFA (Unique KC)} & \textbf{SPARFA-Lite} & \textbf{Tensor Factorization} \\ 
\midrule
Lesson 1 (M) & 0.4790 & 0.4331 & 0.4733 & 0.4550 & 0.6287 & \textbf{0.4328} \\ 
Lesson 1 (H) & 0.4178 & 0.4011 & 0.4315 & 0.4152 & 0.4850 & \textbf{0.3973} \\ 
Lesson 2 (M) & 0.4575 & 0.3957 & 0.4557 & 0.4390 & 0.5691 & \textbf{0.3840} \\ 
Lesson 2 (E) & 0.4261 & 0.4118 & 0.4415 & 0.4389 & 0.4969 & \textbf{0.4085} \\ 
Lesson 2 (H) & 0.4512 & 0.3861 & 0.4490 & 0.4358 & 0.5449 & \textbf{0.3715} \\
Lesson 3 (M) & 0.4077 & 0.3936 & 0.4227 & 0.4098 & 0.4644 & \textbf{0.3911}  \\
Lesson 3 (H) & 0.4768 & 0.4004 & 0.4782 & 0.4704 & 0.5826 & \textbf{0.3948} \\
Lesson 4 (M) & 0.4766 & 0.4583 & 0.4775 & 0.4799 & 0.6008 & \textbf{0.4438} \\
Lesson 4 (E) & 0.4929 & 0.4750 & 0.4989 & 0.4939 & 0.5789 & \textbf{0.4457} \\
Lesson 4 (H) & 0.4968 & 0.4605 & 0.5154 & 0.5027 & 0.6544 & \textbf{0.4545} \\ 
\bottomrule
\end{tabularx}
\end{table*}

\begin{table*}[ht!]
\caption{Comparative analysis of model prediction accuracy in AutoTutor ARC lesson dataset using \(MAE\). Bold values represent the lowest values.}\label{tbl_mae}
\centering
\begin{tabularx}{\textwidth}{@{}p{1.9cm} >{\centering\arraybackslash}p{2.2cm} >{\centering\arraybackslash}p{2.3cm} >{\centering\arraybackslash}p{2.2cm} >{\centering\arraybackslash}p{2.2cm} >{\centering\arraybackslash}p{2cm} >{\centering\arraybackslash}p{2.6cm}@{}}
\toprule
\textbf{Dataset} & \textbf{BKT (Single KC)} & \textbf{BKT (Unique KC)} & \textbf{PFA (Single KC)} & \textbf{PFA (Unique KC)} & \textbf{SPARFA-Lite} & \textbf{Tensor Factorization} \\ 
\midrule
Lesson 1 (M) & 0.4623 & \textbf{0.3777} & 0.4500 & 0.4071 & 0.3954 & 0.3806 \\ 
Lesson 1 (H) & 0.3535 & 0.3347 & 0.3340 & 0.3350 & \textbf{0.2387} & 0.3178 \\ 
Lesson 2 (M) & 0.4252 & 0.3257 & 0.4150 & 0.3806 & 0.3245 & \textbf{0.2740} \\ 
Lesson 2 (E) & 0.3731 & 0.3461 & 0.3666 & 0.3507 & \textbf{0.2503} & 0.3052 \\ 
Lesson 2 (H) & 0.4184 & 0.3141 & 0.4003 & 0.3595 & 0.2979 & \textbf{0.2904} \\
Lesson 3 (M) & 0.3330 & 0.3175 & 0.3293 & 0.3256 & \textbf{0.2158} & 0.2805 \\
Lesson 3 (H) & 0.4588 & 0.3275 & 0.4446 & 0.4380 & 0.3400 & \textbf{0.3030} \\
Lesson 4 (M) & 0.4598 & 0.4106 & 0.4437 & 0.4487 & 0.3625 & \textbf{0.3494} \\
Lesson 4 (E) & 0.4819 & 0.4304 & 0.4789 & 0.4806 & 0.3583 & \textbf{0.3372} \\
Lesson 4 (H) & 0.4945 & 0.3998 & 0.4914 & 0.4815 & 0.4287 & \textbf{0.3832} \\ 
\bottomrule
\end{tabularx}
\end{table*}

Here, to answer the \textbf{RQ1}, we present the results about the sparsity levels for the learning performance data and the latent feature \(K\) obtained from the Tensor Factorization computing process for various lessons and their corresponding difficulty levels. This analysis aims to understand the extent of missing data and to determine the number of latent features required to accurately represent the underlying patterns. Table~\ref{tb:sparse} presents the measured sparsity levels and the number of latent features \(K\) derived via tensor factorization. The sparsity levels across the dataset predominately fall within the \(80\%\) to \(90\%\) range, except for Lesson 4 (E), which stands at \(64.58\%\). Higher sparsity levels indicate a greater proportion of missing values in the dataset, and vice versa. The values of latent features \(K\) exhibit variation across different lessons and their corresponding difficulty levels, typically requiring between 4 to 7 latent features to accurately represent the underlying patterns in the learning performance data. 

\begin{table}[ht!]
\caption{The sparsity levels and obtained latent features by tensor factorization.}\label{tb:sparse}
\centering
\begin{tabularx}{\columnwidth}{@{}l >{\centering\arraybackslash}X >{\centering\arraybackslash}X@{}}
\toprule
\textbf{Dataset} & \textbf{Sparsity Levels} & \textbf{\(K\) (Latent Features)} \\ 
\midrule
Lesson 1 (M) & 84.94\% & 6 \\ 
Lesson 1 (H) & 89.89\% & 7 \\ 
Lesson 2 (M) & 84.02\% & 6 \\ 
Lesson 2 (E) & 81.25\% & 4 \\ 
Lesson 2 (H) & 85.45\% & 6 \\
Lesson 3 (M) & 81.25\% & 5 \\
Lesson 3 (H) & 82.04\% & 6 \\
Lesson 4 (M) & 86.27\% & 6 \\
Lesson 4 (E) & 64.58\% & 6 \\
Lesson 4 (H) & 86.57\% & 5 \\
\bottomrule
\end{tabularx}

\vspace{0.1cm} 

\noindent\justifying\textbf{Notes:}
The table summarizes the sparsity levels and the number of latent features (\(K\)) obtained from tensor factorization across various datasets. The datasets are divided by lesson and categorized as M (Medium), H (Hard), and E (Easy). Higher sparsity levels indicate a greater proportion of missing values in the dataset. The number of latent features \(K\) represents the dimensionality of the latent space resulting from the factorization process.
\end{table}

\subsection{Predictive Accuracy for Data Imputation}

As demonstrated by the predictive accuracy results based on the two metrics, \textit{RMSE} and \textit{MAE}, which address \textbf{RQ1}, the following analysis provides insights into model performance. 

Table~\ref{tbl_rmse} demonstrates that Tensor Factorization consistently achieves the lowest \textit{RMSE} values across all lessons and difficulty levels (medium, easy, and hard), highlighting its superior prediction accuracy (with the lowest \textit{RMSE} values bolded in the table). This indicates that Tensor Factorization is particularly effective in predicting performance compared to the other three KT models, including BKT, PFA, and SPARFA-Lite. The superior performance of Tensor Factorization in learner modeling also reflects its accuracy in predicting missing data through its mathematical imputation methods, addressing a critical need for data imputation in adult literacy. Among the \textit{RMSE} values of Tensor Factorization across all lessons and difficulty levels, the model achieves the lowest \textit{RMSE} in Lesson 2 (H) with 0.3715 and the highest \textit{RMSE} in Lesson 4 (H) with 0.4545, indicating that its prediction accuracy is best in Lesson 2 (H) and less effective in Lesson 4 (H). Additionally, the \textit{RMSE} from the BKT (unique KC) model is quite close to Tensor Factorization and outperforms other models in several instances. For example, in Lesson 1 (M), BKT (unique KC) achieves an \textit{RMSE} of 0.4331 compared to Tensor Factorization's 0.4328, and in Lesson 3 (H), BKT (unique KC) achieves an \textit{RMSE} of 0.4004 compared to Tensor Factorization's 0.3948. This trend is also observed in other lessons, including Lesson 1 (H), Lesson 2 (M), Lesson 2 (E), Lesson 2 (H), Lesson 3 (M), Lesson 4 (M), Lesson 4 (E), and Lesson 4 (H). These trends reflect the relatively higher accuracy of BKT compared to the other models, except for Tensor Factorization. 

Table~\ref{tbl_mae} presents a comparison of \textit{MAE} values across various models and lessons, with the lowest values highlighted in bold. Notably, Tensor Factorization frequently achieves the lowest \textit{MAE} values, signaling its robustness in closely estimating actual learning performance. Nonetheless, some models such as SPARFA-Lite exhibit exceptional performance in specific lessons including Lesson 1 (H) with an \textit{MAE} of 0.2387, Lesson 2 (E) with an \textit{MAE} of 0.2503, and Lesson 3 (M) with an \textit{MAE} of 0.2158, with BKT (Unique KC) also excelling in Lesson 1 (M) with an \textit{MAE} of 0.3777, indicating their potential effectiveness under specific conditions. 

To determine if the \textit{MAE} values obtained by Tensor Factorization are significantly lower than those from other models, an ANOVA analysis was conducted using \textit{MAE} as the dependent variable and a binary factor indicating Tensor Factorization as the independent variable. During the ANOVA analysis, the two degrees of freedom are as follows: \(df_{between}=5\) for between groups, and \(df_{within}=54\) for within groups. The \textit{F-value} is calculated to be 7.2003, with a corresponding \textit{p-value} of \(3.164 \times 10^{-5}\), which is less than 0.05. This indicates statistically significant differences in \textit{MAE} between Tensor Factorization and the other models. 

In conclusion, Tensor Factorization allows for highly accurate prediction of missing performance data based on the existing data distribution, enabling a comprehensive dataset for data imputation. 

\subsection{Identification of Individualized Learning Performance Patterns} 

To address \textbf{RQ2}, this section presents the results of identifying individualized learning performance patterns through clustering analysis. By leveraging a dense 3-dimensional tensor representation of data imputed from Tensor Factorization, we analyze distinct learning performance clusters. 

\begin{figure*}[ht!]
    \centering
    \begin{subfigure}[b]{0.48\textwidth}
        \centering    \includegraphics[width=\textwidth]{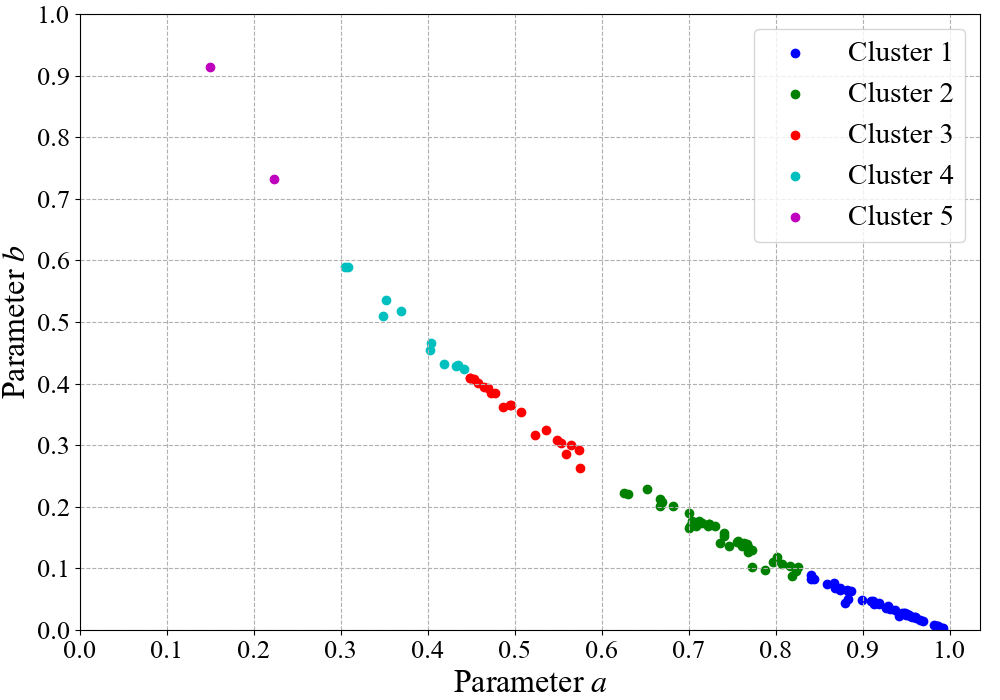}
        \caption{Distribution of parameter \(a\) and \(b\).}
        \label{fig:cluster_a}
    \end{subfigure}
    \hfill
    \begin{subfigure}[b]{0.48\textwidth}
        \centering
        \includegraphics[width=\textwidth]{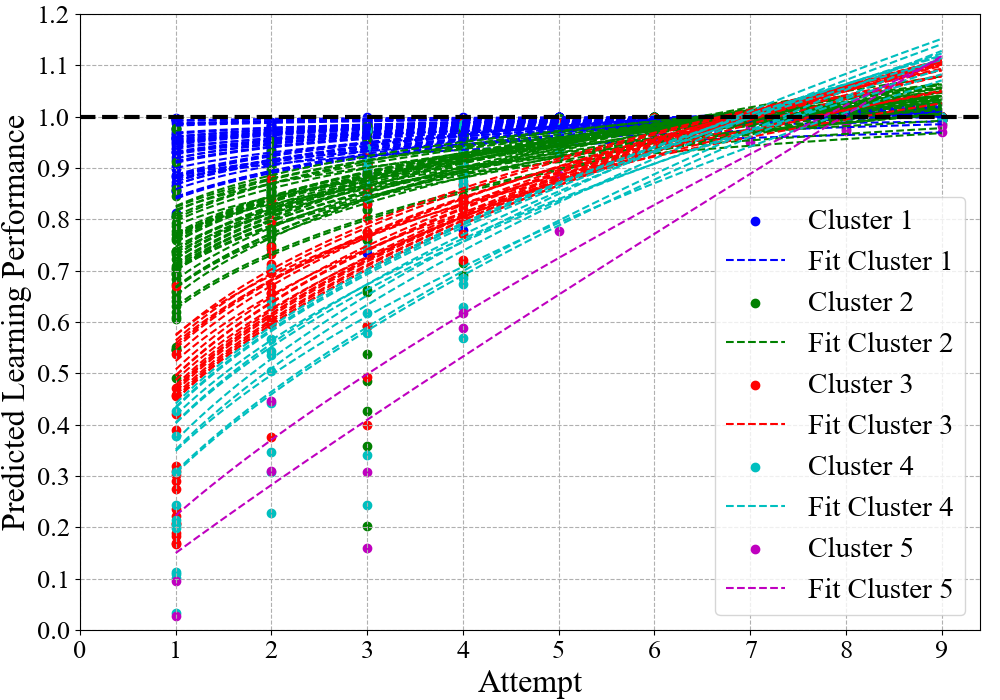}
        \caption{Curve fitting of learning performance along attempt}
        \label{fig:cluster_b}
    \end{subfigure}
    \caption{Estimates of parameter estimates \(a\) and \(b\) and identification of learning performance patterns through clustering.}
    \label{fig:parameter_evaluation}
\end{figure*}

Fig.~\ref{fig:parameter_evaluation} illustrates distinct learning performance patterns identified through clustering, based on a dense 3D tensor representation of data imputed from Tensor Factorization, using the 6th question of Lesson 2 (M) as a case study. It identifies five distinct clusters based on fluctuations in parameter \(a\) and \(b\) from power-law function fitting on the learners-attempts matrix data (as mentioned before) for the 6th question, each depicted in different colors as shown in Fig.~\ref{fig:cluster_a}. From Cluster 1 to Cluster 5, the \(a\) value decreases, starting largest in Cluster 1 and gradually reducing through to Cluster 5. Conversely, the \(b\) value increases across the clusters from Cluster 1 to Cluster 5, indicating an inverse relationship between \(a\) and \(b\) among different clusters. 

Fig.~\ref{fig:cluster_b} displays the clustering results using identical color coding for the five clusters and shows that predicted learning performance (probability-based) generally increases monotonically with the number of attempts, approaching or reaching a performance level of around 1 (\textit{y}-axis value). This trend persists across all clusters, although at varying rates. Specifically, most fitting curves approach around 1 within seven attempts, indicating that the majority of learners likely do not need more than seven attempts to achieve theoretical complete mastery, a pattern also observed in other cases not shown. Note that the fitting curves express the overall trend, while scatter points represent the actual performance data. 

\subsection{Results of Scalable Data Augmentation}

This subsection presents the data augmentation results based on GenAI models with scalable sampling, addressing \textbf{RQ2}. The analysis highlights how these models effectively enhance data diversity and volume, thereby boosting the robustness of subsequent machine learning applications. Furthermore, a detailed comparison between Vanilla GAN and GPT-4o based augmentation methods highlights which approach offers greater stability across varying sample sizes. 

\begin{figure*}[ht!]
    \centering
    \includegraphics[width=\textwidth]{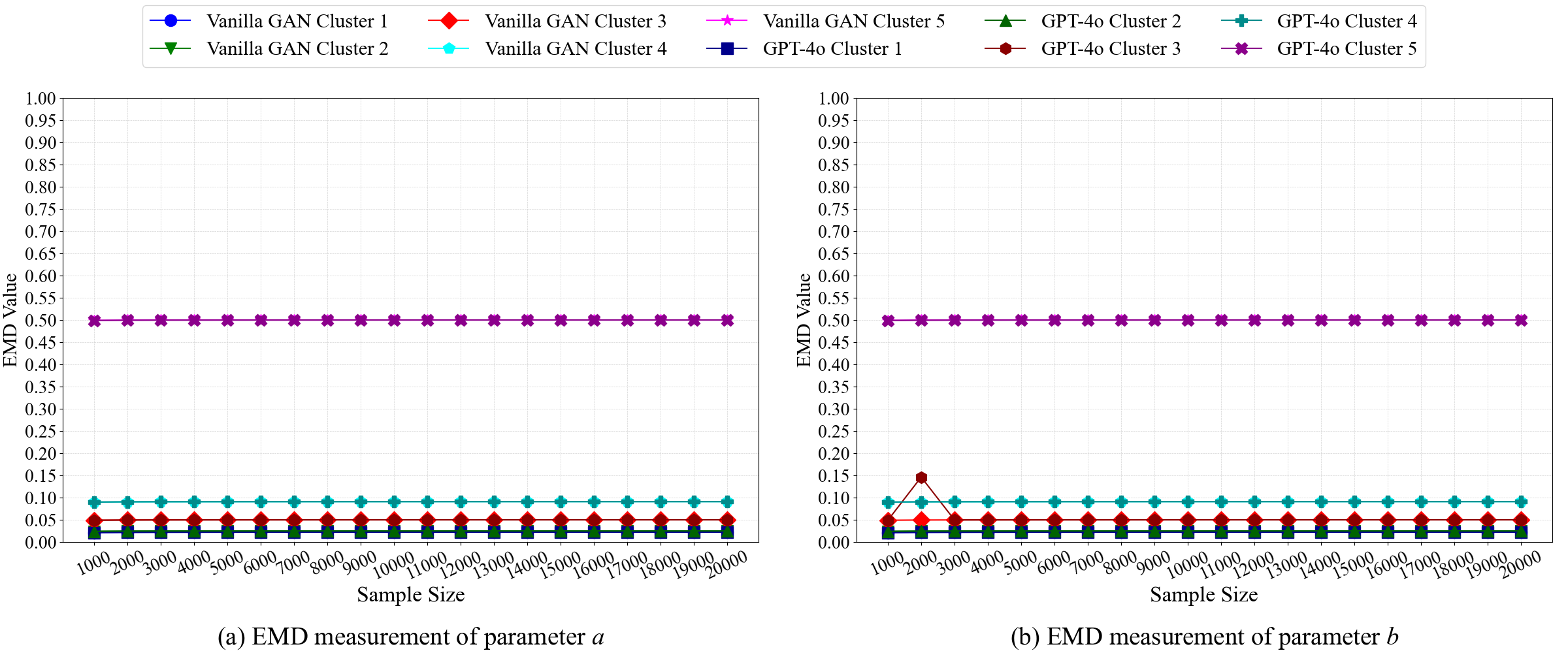}
    \caption{\textit{EMD} measurement for scalable sampling by data augmentation. The original sample size is 20, and the augmentations are shown in increments of 1,000, with total sizes ranging from 1,000 to 20,000.}
    \label{fig:Parameter_a_b_emd_comparison}
\end{figure*}

\begin{figure*}[h!t]
\centering
\includegraphics[width=7.0in]{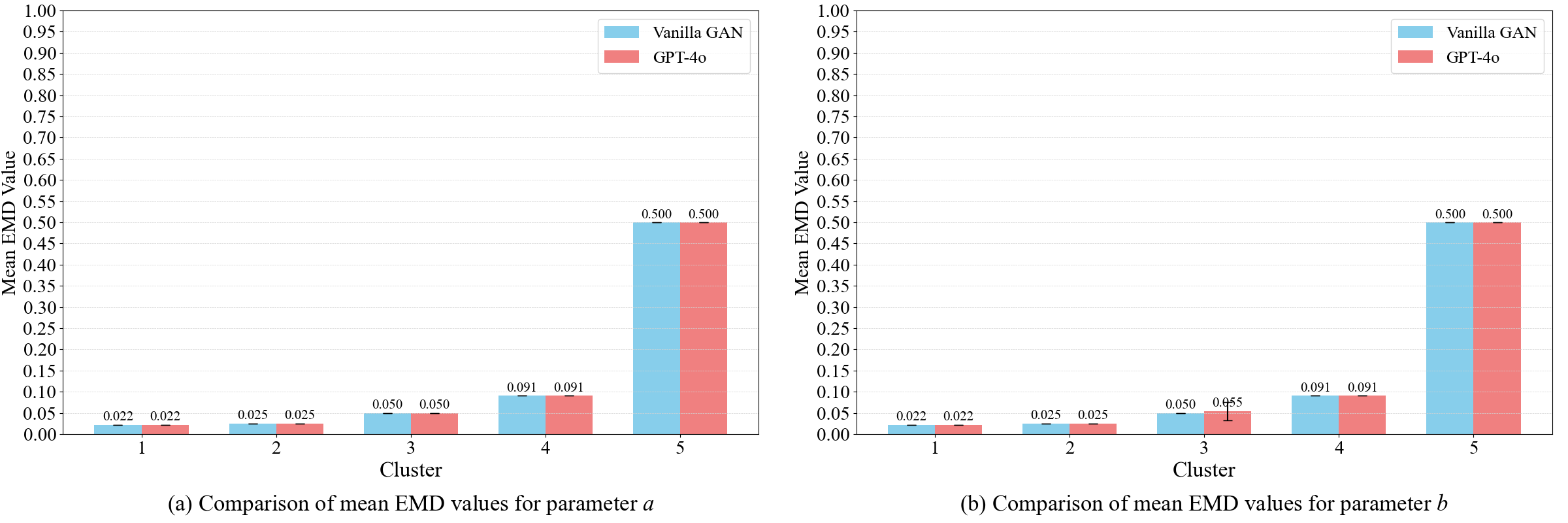}
\caption{Comparison of \textit{EMD} measurement for data augmentation between Vanilla GAN and GPT-4o.}
\label{fig:Parameter_a_b_mean_emd_bar_comparison}
\end{figure*}

Fig.~\ref{fig:Parameter_a_b_emd_comparison} illustrates the \textit{EMD} measurement results for divergence in power-law fitting parameters, comparing augmented learning performance data to original data. It covers the full scope of parameter \(a\) and \(b\) across five clusters for the example \(6^{th}\) question of Lesson 2(M). In both models (Vanilla GAN and GPT-4o), the parameters \(a\) and \(b\) generally exhibit stable and relatively low \textit{EMD} (mostly below 0.10) across increasing sample sizes (from 1,000 to 20,000) for most clusters. However, exceptions are observed in GPT-4o Cluster 5, where the EMD remains around 0.5 for both parameters \(a\) and \(b\) across all sample sizes. This generally demonstrates that both Vanilla GAN and GPT-4o models can generate stable augmented data by effectively learning the original data distributions, even as the sample size scales. Fig.~\ref{fig:Parameter_a_b_mean_emd_bar_comparison} presents the mean \textit{EMD} values across all clusters. The mean EMD values for Vanilla GAN and GPT-4o across both parameters are nearly identical, except for parameter \(b\) in Cluster 3, where Vanilla GAN has a slightly lower value (0.050 compared to GPT's 0.055). The overall mean EMD values across all six clusters are 0.137 for Vanilla GAN and 0.138 for GPT-4o, indicating closely aligned performance between the two generative models. This similarity suggests that both Vanilla GAN and GPT-4o exhibit comparable accuracy in capturing the data distributions across clusters, with minimal divergence between them. In the exceptional case of GPT-4o Cluster 5, however, we hypothesize that the observed differences stem from the model's challenge in capturing this cluster's specific data distribution, likely due to its high sparsity with data from only two users, as shown in Fig.~\ref{fig:parameter_evaluation}). In conclusion, while both Vanilla GAN and GPT-4o models generally produce augmented data that aligns closely with the original data across varying conditions, including quantified learning features, parameters \(a\) and \(b\), and expanded sample sizes, Vanilla GAN demonstrates slightly better performance in certain clusters, specifically the Cluster 3 for parameter \(b\), where its where its EMD values are marginally lower. Overall, both models exhibit stable and reliable performance, with Vanilla GAN performing slightly better than GPT-4o in certain scenarios, and minimal divergence from the original data distributions in most cases. 

\begin{figure*}[ht!]
    \centering
    \begin{minipage}{0.49\textwidth}
        \centering
\includegraphics[width=\textwidth]{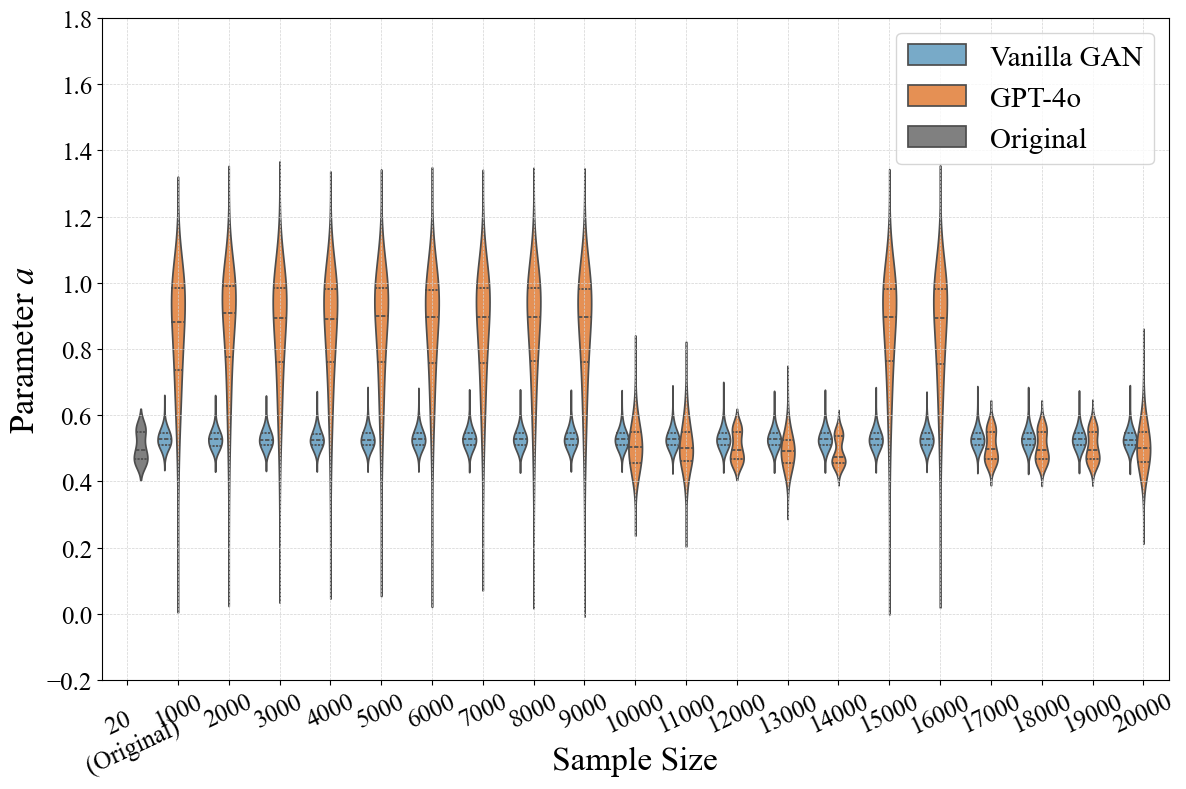}
    \subcaption{Distribution of parameter \(a\)}
        \label{fig:parameter_a}
    \end{minipage}
    \hfill
    \begin{minipage}{0.49\textwidth}
        \centering
\includegraphics[width=\textwidth]{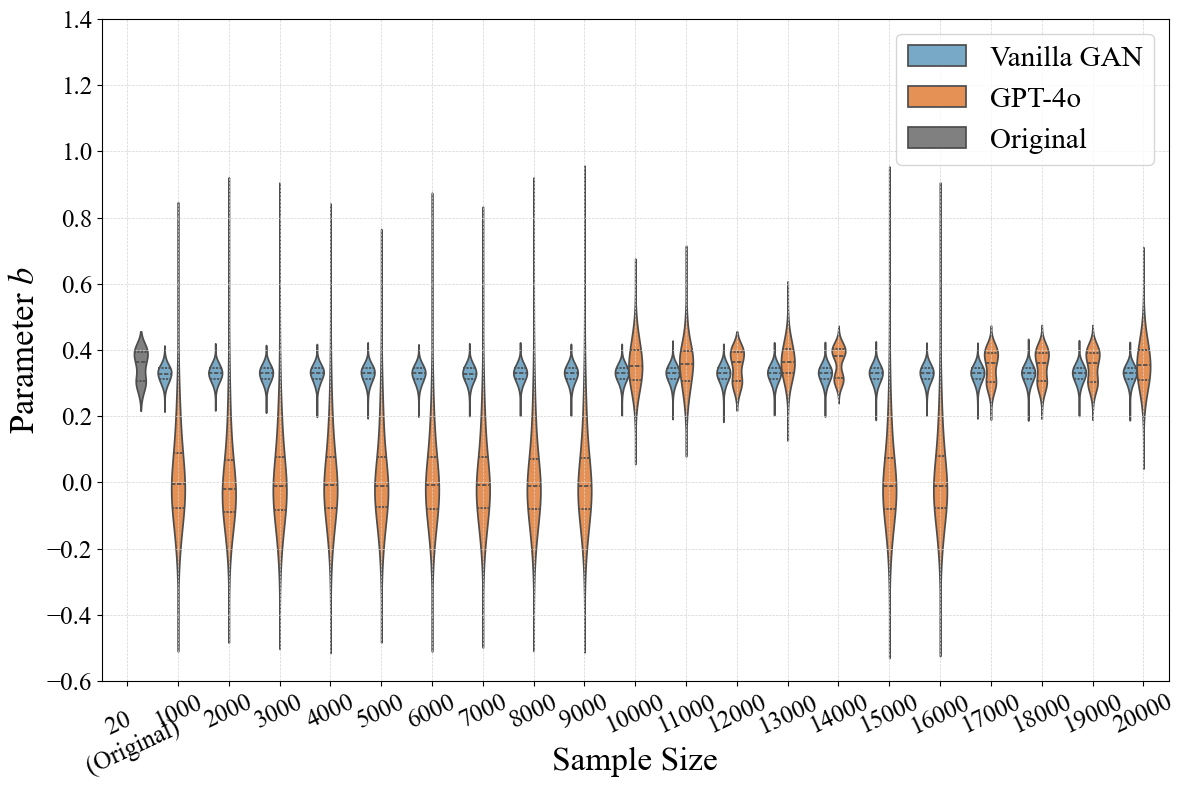}
    \subcaption{Distribution of parameter \(b\)}
        \label{fig:parameter_b}
    \end{minipage}    \caption{Distributions of parameters in augmented learning performance data.}
\label{fig:parameter_augmentation}
\end{figure*} 

\begin{figure*}[ht!]
    \centering
    \begin{minipage}{0.49\textwidth}
        \centering
\includegraphics[width=\textwidth]{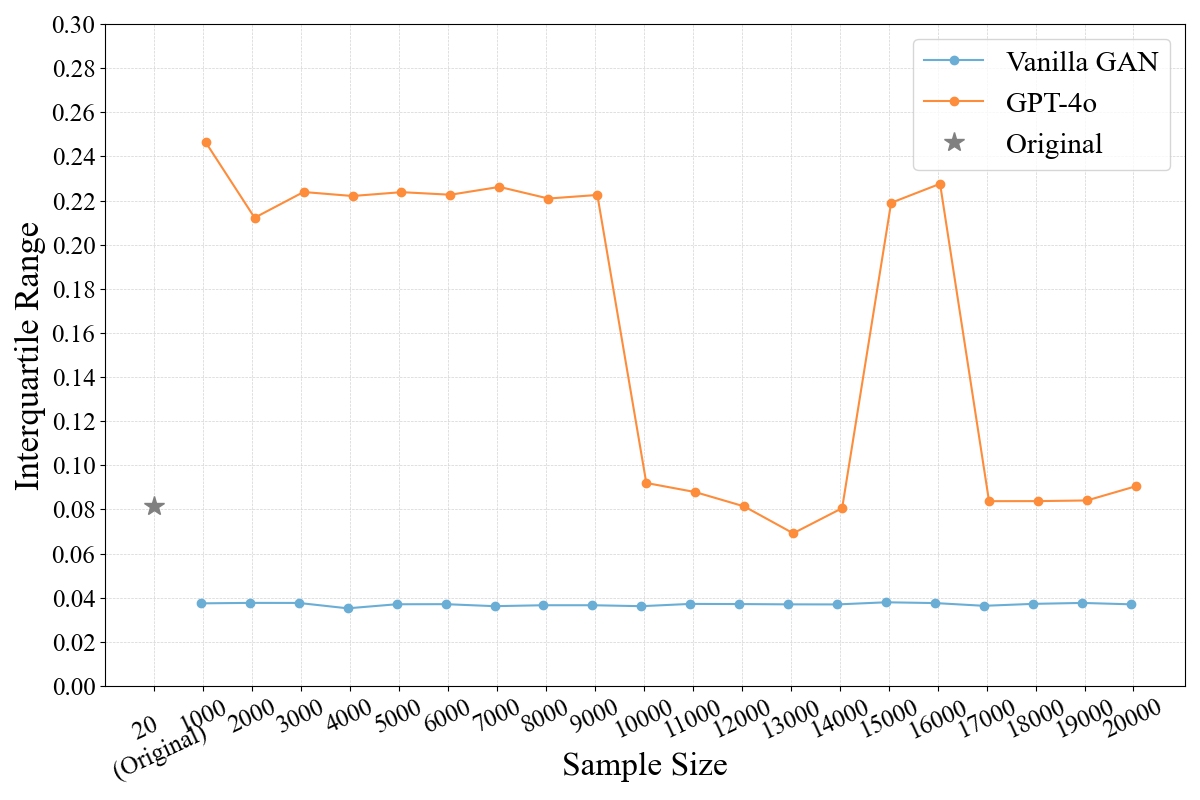}
    \subcaption{Interquartile range of parameter \(a\)}
        \label{fig:parameter_a_iqr}
    \end{minipage}
    \hfill
    \begin{minipage}{0.49\textwidth}
        \centering
\includegraphics[width=\textwidth]{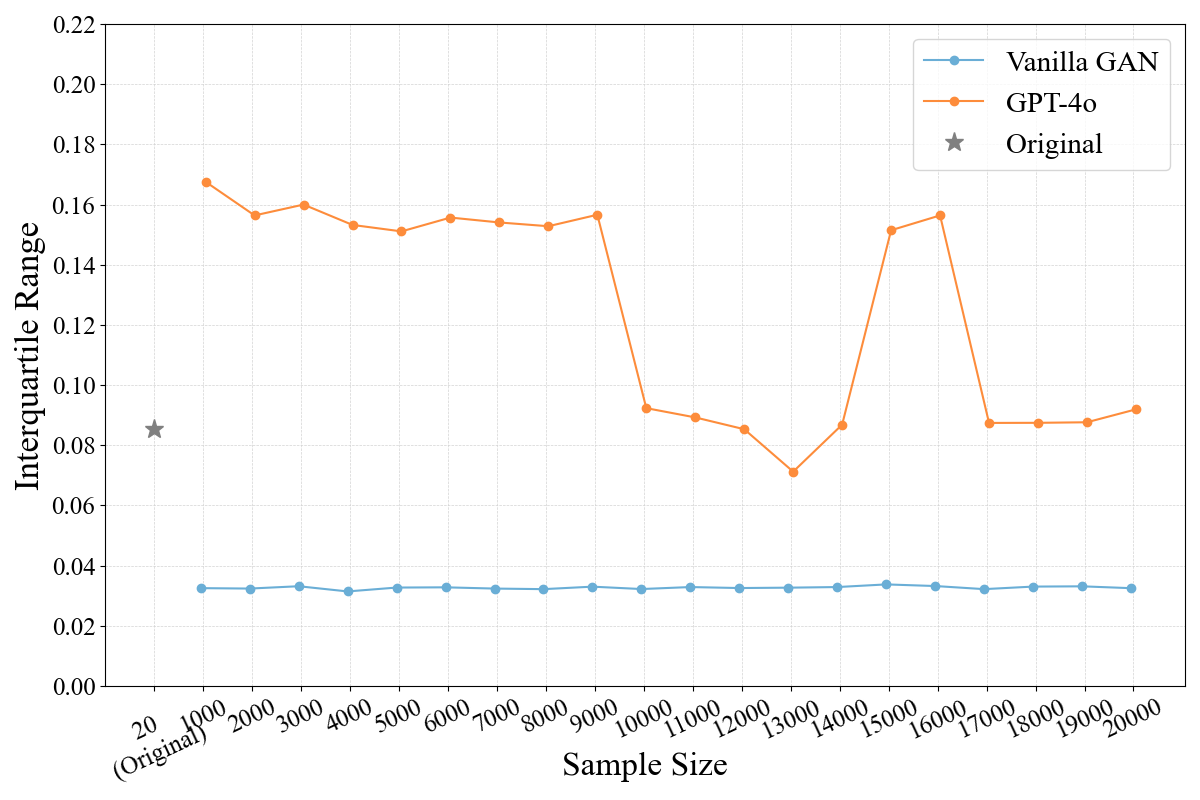}
    \subcaption{Interquartile range of parameter \(b\)}
        \label{fig:parameter_b_iqr}
    \end{minipage}
    \caption{Variance comparison of parameters in augmented learning performance data.}
\label{fig:parameter_IQR}
\end{figure*}

\begin{figure*}[ht!]
    \centering
    \begin{minipage}{0.49\textwidth}
        \centering
\includegraphics[width=\textwidth]{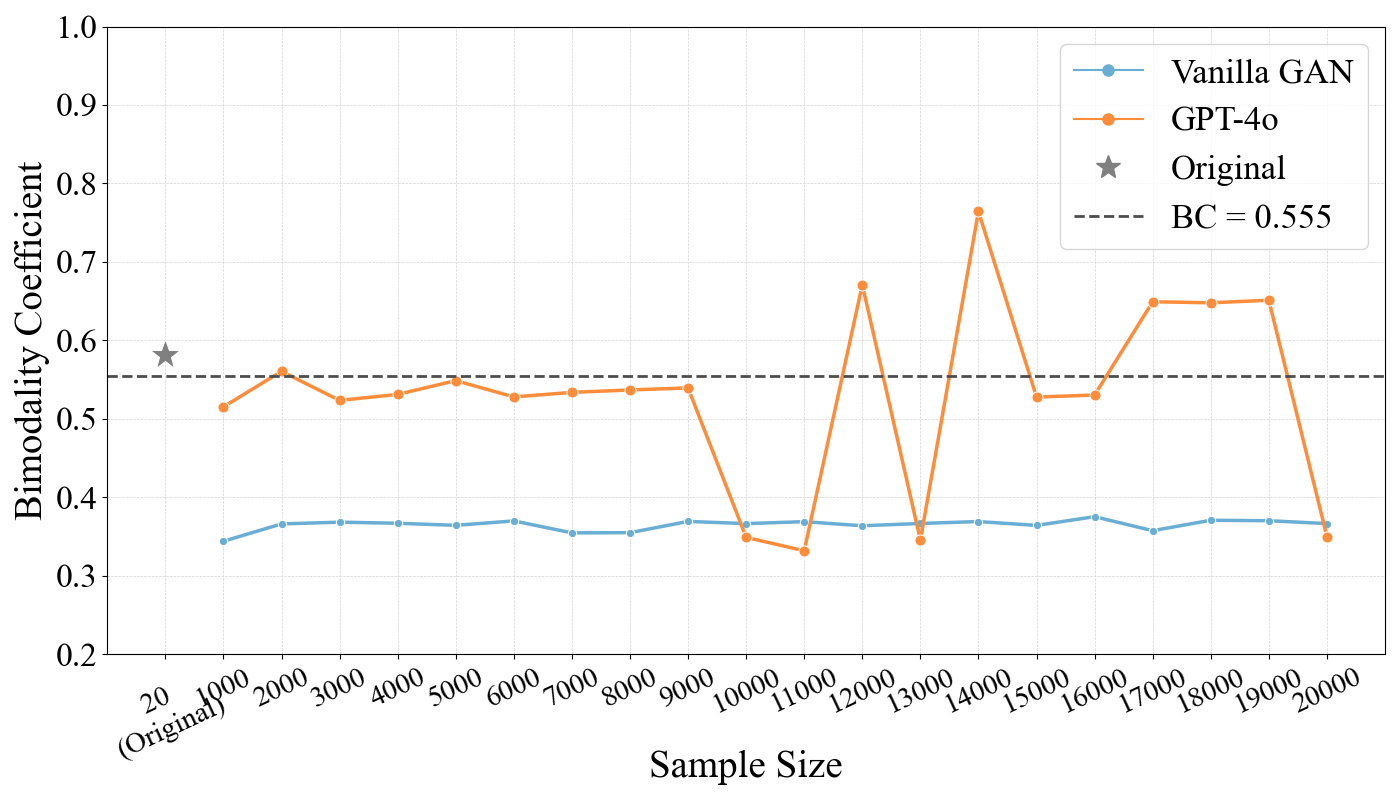}
    \subcaption{Bimodality coefficient of parameter \(a\)}
        \label{fig:parameter_a_bio}
    \end{minipage}
    \hfill
    \begin{minipage}{0.49\textwidth}
        \centering
\includegraphics[width=\textwidth]{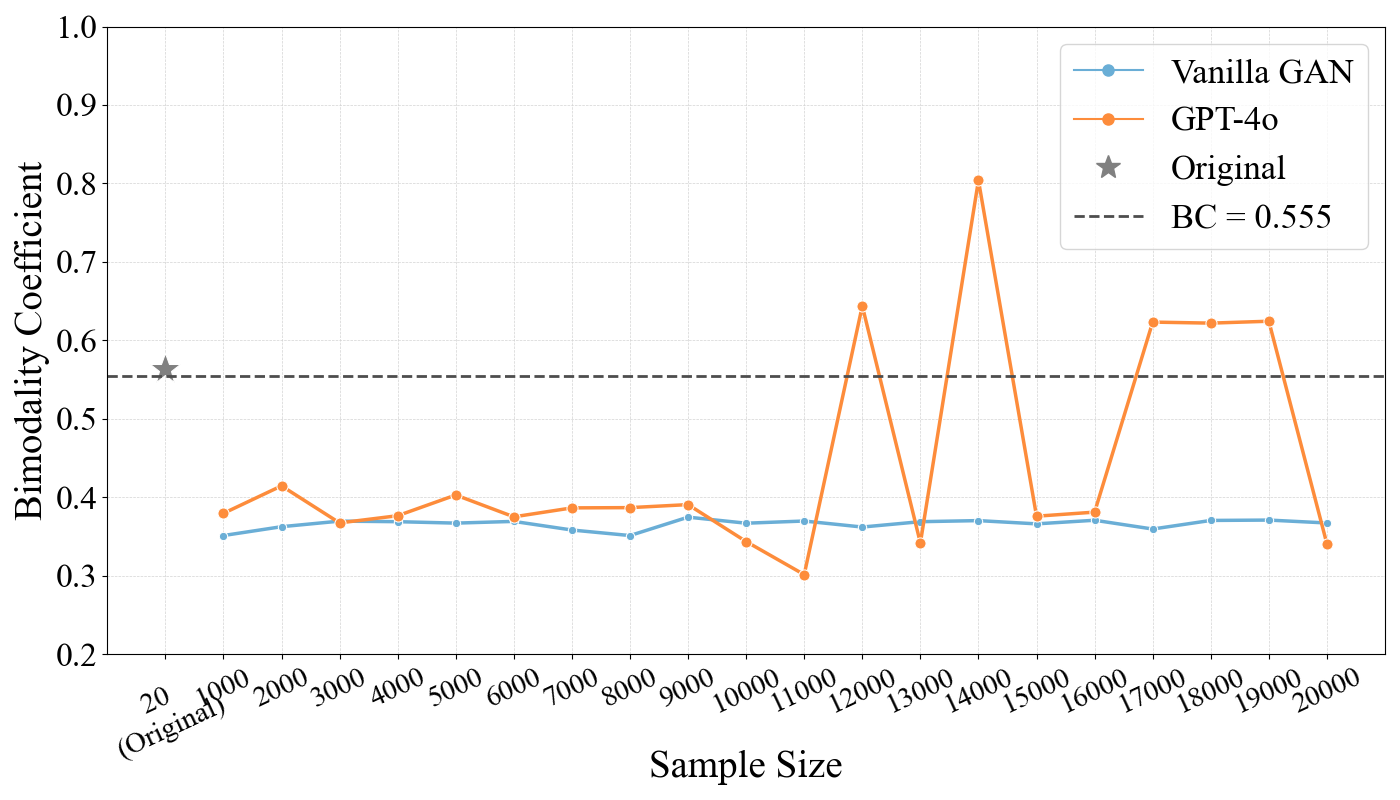}
    \subcaption{Bimodality coefficient of parameter \(b\)}
        \label{fig:parameter_b_bio}
    \end{minipage}
    \caption{Comparison of the bimodality coefficient of parameters in augmented learning performance data.}
\label{fig:parameter_biomodality}
\end{figure*}

To further investigate these observations, we now examine the detailed distribution patterns of parameters \(a\) and \(b\) obtained from the augmented datasets, as illustrated in Fig.~\ref{fig:parameter_augmentation}. This figure presents the unnormalized distributions of parameters \(a\) and \(b\) obtained through power-law fitting on the augmented learning performance dataset, using both Vanilla GAN and GPT-4o based augmentation. For clarity, Cluster 3 of the 6th question from Lesson 2 (M) is used as an example to highlight the differences in greater detail. By focusing on unnormalized data, we aim to preserve the original scale of the parameters and directly compare their raw distributions. Note that the grey color represents the original data distribution. Moreover, the IQR, depicted in Fig.~\ref{fig:parameter_IQR}, supplements this analysis by providing a refined measure of distribution spread. The combined insights from these two figures are summarized in the following paragraphs. 

Fig.~\ref{fig:parameter_a} illustrates the distribution of parameter \(a\) for Vanilla GAN and GPT-4o models. The following observations can be made. First, the parameter distribution of \(a\) obtained from Vanilla GAN-based augmentation generally falls within the range of the original data, with a slightly extended tail beyond the original maximum value. Compared to the original distribution, the Vanilla GAN-based distributions maintain a similar shape with a concentrated and narrower central region in the violin plots, indicating that the Vanilla GAN model effectively captures the core characteristics of the original data while introducing minimal variability. The IQR values for Vanilla GAN, as shown in Fig.~\ref{fig:parameter_a_iqr}, remain consistently stable with minimal fluctuations across all sample sizes, reflecting reliable behavior and closely aligning with the original data’s variability. This consistent performance in terms of parameter \(a\) highlights the stability and dependability of Vanilla GAN-based augmentation in preserving the structure and core characteristics of the original distribution. In contrast, the GPT-4o-based augmentation exhibits considerably broader parameter \(a\) distributions for most sample sizes, suggesting higher variability. This variability is particularly pronounced for certain sample sizes, such as 1,000 to 9,000, and 15,000 to 16,000, where the GPT-4o distributions display elongated tails and exhibit distinctly different central regions. Compared to the original distribution, these central regions are not only wider but also shifted, indicating significant deviations in both dispersion and centrality from the original data's core characteristics. This higher variability in parameter \(a\) is further evidenced by the fluctuating IQR values for GPT-4o-based augmentation in Fig.~\ref{fig:parameter_a_iqr}. However, for specific sample sizes, including 10,000, 12,000, 13,000, 14,000, and 17,000 to 20,000, the GPT-4o-based distributions exhibit relatively smaller ranges closer to the original data, which is also reflected in lower IQR values for these cases, though they still extend beyond the original distribution. These observations show that GPT-4o-based augmentation has high variability and inconsistent performance in approximating the original distribution, capturing its core characteristics only occasionally. In contrast, Vanilla GAN-based augmentation exhibits consistently lower variability, with stable IQR values, highlighting its reliability in preserving the original data's structure and variability.

Fig.~\ref{fig:parameter_b} shows the distribution of parameter \(b\) for Vanilla GAN-based augmentation. First, the parameter distribution of obtained from Vanilla GAN-based augmentation also generally aligns with the original data, particularly in terms of maximum and minimum values. The distributions exhibit a similar shape, with slightly extended tails beyond the original range but no significant deviations. The IQR values for Vanilla GAN, as shown in Fig.~\ref{fig:parameter_b_iqr}, remain consistently stable across all sample sizes, reflecting Vanilla GAN's stable and reliable performance in data augmentation. This consistent behavior in parameter \(b\) highlights the Vanilla GAN model's effectiveness in preserving the structure and variability of the original distribution with minimal distortion. In contrast, the GPT-4o-based augmentation exhibits significantly higher variability for parameter \(b\), as evidenced by the broader distributions and elongated tails across most sample sizes, as shown in Fig.~\ref{fig:parameter_b}. The GPT-4o distributions display marked deviations from the original data, particularly for sample sizes between 1,000 and 9,000 and again at 15,000 to 16,000, where the tails become elongated. The dispersion is further reflected in the fluctuating IQR values for GPT-4o, as seen in Fig.~\ref{fig:parameter_b_iqr}. However, there is occasional alignment with the original data for specific sample sizes, such as 10,000 to 14,000 and 17,000 to 20,000. In these cases, the distributions are relatively closer to the original data compared to other sample sizes, as observed in both Fig.~\ref{fig:parameter_b} and Fig.~\ref{fig:parameter_b_iqr}. These observations indicate that, for parameter \(b\), Vanilla GAN-based augmentation demonstrates consistent and reliable performance, effectively capturing the original distribution’s structure and variability. In contrast, GPT-4o-based augmentation exhibits greater variability and inconsistency, with fluctuations in alignment and dispersion that make it less reliable for preserving the original distribution’s characteristics. 

As indicated by the distribution shape, we analyze the parameters \(a\) and \(b\) through BC, which provides a quantitative measure of bimodality, as illustrated in Fig.~\ref{fig:parameter_biomodality}. The threshold \(BC=0.555\), marked in the Fig.~\ref{fig:parameter_biomodality}, indicates that values exceeding this threshold generally signify bimodality, while lower values suggest unimodality. The original data distribution for Cluster 3 exhibits bimodality for both parameter \(a\) and \(b\), as evidenced by BC values greater than 0.555, represented by the dark dot labeled ``Original'' in the figure. As shown in Fig.~\ref{fig:parameter_biomodality}, the Vanilla GAN-generated data predominantly exhibits unimodality across all sample sizes for both parameters, with BC values remaining below the 0.555 threshold. On the other hand, the GPT-4o-generated data generally exhibits unimodality, although it transitions to bimodality in specific cases. Notably, unimodality is observed at sample sizes of 12,000, 14,000, 17,000, 18,000, and 19,000 for both parameters \(a\) and \(b\). The Vanilla GAN show stable though that distribution shape are different original data, the GPT-4o is high variant but some cases distribution shape more closer to original data distributions shape. These findings indicate that the Vanilla GAN model consistently generates unimodal distributions, which, while stable, deviate from the original data's bimodal characteristics. In contrast, the GPT-4o model exhibits greater variability in its distributions. In specific instances, the distributions generated by GPT-4o align more closely with the original data's bimodal structure, suggesting that GPT-4o has the potential to better capture certain features of the original data under specific conditions. 

\section{Discussion}

This study proposes a systematic augmentation framework integrating multidimensional learner modeling with generative AI models, specifically Vanilla GANs and GPT-4o, to address the critical issue of data sparsity in ITSs. Our primary objective was to enhance the quality and scalability of learning performance data in adult reading comprehension using tensor factorization for data imputation and generative models for data augmentation. Guided by the research questions on the effectiveness of tensor factorization for imputing student performance data and the utilization of GenAI models for tailored data augmentation, our findings reveal that tensor factorization outperforms traditional modeling methods, providing higher predictive accuracy and more effective handling of missing values. Building on the densified 3D tensor-based learning performance data, individual learning performance patterns were identified through clustering, and both Vanilla GAN and GPT-4o models were employed to augment learning performance data tailored to these patterns. Results demonstrate that augmented datasets from both Vanilla GAN and GPT-4o align closely with the original data distribution, as evidenced by divergence measurements using the EMD. However, key differences between the models were observed: Vanilla GAN-generated data consistently produces stable, unimodal distributions with lower variability and divergence, whereas GPT-4o generated data exhibits higher variability and occasional bimodality, aligning more closely with the original data’s structure under specific sample sizes. These complementary strengths underscore Vanilla GAN's reliability and GPT-4o's potential to replicate complex patterns, highlighting their utilities for enhancing learning performance data. 

High sparsity levels in the dataset, as highlighted in Table \ref{tb:sparse}, significantly impact modeling and analytics, leading to challenges such as modeling bias, reduced performance, and difficulties in knowledge tracing \cite{saarela2017automatic,lee2022contrastive}. Other research has also highlighted similar high sparsity levels in educational datasets. For example, Saarela's research \cite{saarela2017automatic} on student performance data notes sparsity levels ranging from 30\% to 74\% in the PISA 2012 dataset, and Thai-Nghe's study \cite{nguyen2011predicting} on the Algebra dataset reports extreme sparsity, with rates reaching up to 99.81\%. The latent features captured through tensor factorization quantitatively reflect learner-specific characteristics, such as learning abilities and personalities. These characteristics of learners remain stable over time in the reading comprehension context. The number of features identified, averaging around six, closely aligns with Graesser-McNamara's multi-level theoretical framework \cite{graesser2011computational} for reading comprehension, which includes six levels: word, explicit textbase, referential situation model, discourse genre and rhetorical structure, and pragmatic communication level \cite{shi2017using}. However, further research is needed to verify this alignment and explore the potential connections. 

Constructing a 3D tensor of learning performance data provides a foundation for analyzing patterns across learners, questions, and attempts. Tensor factorization captures complex interactions among these dimensions and estimates missing values for unattempted questions. This approach builds on successful applications of tensor methods in learning performance analysis, as shown in prior studies \cite{thai2012predicting, doan2019rank, zhang2023exploring}. By integrating this modeling technique, we can more accurately impute missing data, thereby enhancing the robustness of the analysis and providing deeper insights into learner behavior and performance trends in ITSs. 

The identification of different learning performance patterns across various clusters highlights the importance of variations in initial learning performance and learning rate as key differentiators among learners. The observed inverse relationship between these two quantified features (\(a\) and \(b\)) reveals a notable learning phenomenon: learners with higher initial abilities are closer to complete mastery and exhibit slower knowledge acquisition, while those with lower initial abilities achieve faster knowledge gains due to their lower starting points. While other past work has found such aptitude-treatment interactions, several other data sets, broader evidence finds that learning rates are largely independent of ability, with exceptions often lying in language learning domains (such as the domain of this study) where individual learning differences can be compensated for less by tailored instructions \cite{liu2017towards}. 

In terms of \textit{EMD} measurements, both Vanilla  GAN and GPT-4o models demonstrate stable performance in data augmentation across varying sample sizes, with Vanilla GAN showing slightly better results than GPT-4o in certain scenarios and exhibiting minimal divergence from the original data distributions in most cases. While Vanilla GAN demonstrates more consistent and stable performance across all conditions, GPT-4o generated data exhibits higher variability and occasional bimodality, allowing it to align more closely with the original data’s structure under specific sample sizes. These observations are further validated through evaluations of the distribution characteristics of parameters \(a\) and \(b\), using visualization techniques (violin plots) and quantitative metrics such as IQR and BC metric. The Vanilla GAN model, with its advanced neural network architecture and thousands of parameters, outperforms traditional statistical methods in capturing the original data distribution, making it a dependable choice for scenarios requiring consistent augmentation across varied conditions. On the other hand, GPT-4o based augmentation closely aligns with the original data distribution, suggesting high fidelity. GPT-4o’s capability to understand context-based numerical values and execute self-driven tasks including self-analysis, self-programming, and self-computing during data augmentation \cite{zhang20243dg} highlights its potential for generating complex data patterns. These capabilities, especially when combined with advanced computational techniques, underscore GPT-4o’s promise for data augmentation tasks requiring nuanced contextual understanding and adaptability. 

When deciding between the two type of models, the choice should be scenario-based and goal-driven, aligning with the specific context and objectives of the task. This study provides a foundational methodology for employing GenAI models, including generative computing models exemplified by GAN and LLMs represented by GPT-4o, while acknowledging that alternative GenAI models could also be applied, with performance varying based on testing and context. A scenario-based approach involves analyzing the context in which the model will be used, such as favoring Vanilla GAN for applications requiring stability and consistency, or GPT-4o for replicating complex, nuanced data patterns. A goal-driven perspective focuses on aligning the model's strengths with the intended outcomes, such as using Vanilla GAN to generate data with minimal variability and bias, or GPT-4o to mimic intricate structures and support diversity. The performance in augmenting learning performance data may vary depending on the specific model chosen, but the complementary strengths of GAN (stability) and GPT-4o (complexity) highlight their utility for different contexts and goals.  

Further, Vanilla GAN-derived parameter estimates exhibited desirable properties for both parameters \(a\) and \(b\): lower variance compared to the original parameter distribution at the cost of small bias in parameter estimates. Lower variance is desirable as it indicates a higher accuracy with which \(a\) can be estimated (in the absence of statistical bias, which was minimal compared to the original parameter distribution). Importantly, this desirable property of Vanilla GAN augmentation is not dependent on the total augmented sample size, suggesting consistent performance irrespective of the degree of data augmentation applied. In contrast, GPT-4o derived parameter estimates demonstrate greater variability compared to the original distribution, indicating less stability in replicating specific parameter characteristics. However, GPT-4o’s high fidelity in preserving the original data’s structural features in certain occasional cases highlights its potential for generating complex and nuanced data patterns. Future research exploring self-driven procedural prompts, such as self-searching, self-programming, and self-computing for numerical value augmentation, may further enhance GPT-4o’s capabilities and improve upon the baseline performance reported in this study. 


The systematic augmentation framework that integrates multidimensional learner modeling through tensor factorization and GenAI models effectively addresses data sparsity issues in ITSs and enables the augmentation of large-scale learning performance data. This approach captures the complexity of human learning performance with greater accuracy than traditional machine learning methods, thereby enhancing predictive capabilities and providing deeper insights into learners' progress. As a result, ITSs can offer more precise recommendations and interventions, leading to personalized and adaptive learning experiences. By generating high-quality augmented data, the framework supports more comprehensive and equitable analysis, reducing biases and enabling more inclusive educational research and practice. This, in turn, helps educators and policymakers make more informed decisions based on robust, large-scale evidence. Additionally, the ability to conduct extensive analyses with augmented data mitigates biases such as group or data selection biases, leading to fairer and more generalizable findings in educational research. Ultimately, this innovative framework significantly contributes to the advancement of AI-driven educational tools, fostering better and more effective instructions and recommendations, and improving educational outcomes for all learners. 

\section{Limitations and Future Works}

This study addresses the significant challenge of data sparsity in ITSs through a systematic augmentation framework that integrates multidimensional learner modeling and GenAI models. However, despite the promising results, there are several limitations that need to be addressed. Firstly, the datasets used in this study were limited to adult reading comprehension cases from the AutoTutor ARC. This restricts the generalizability of our findings to other domains of ITSs, where the characteristics and patterns of learner performance data might differ. Future research should aim to apply this framework to diverse ITSs and learning contexts to validate its applicability and effectiveness. Secondly, the study focused on specific versions of GenAI models, particularly Vanilla GAN and GPT-4o. Our experiments with GPT-4o demonstrated improved performance, particularly in terms of Earth Mover’s Distance (EMD) and visual analyses, compared to GPT-4. However, a more detailed exploration of various GPT versions, such as GPT o-1, is needed to fully understand the specific improvements and differences they offer. Similarly, while we employed the base GAN architecture, Vanilla GAN, for data augmentation, advanced GAN variants like Deep Convolutional GAN (DCGAN), Conditional GAN (cGAN), and Wasserstein GAN (WGAN) could potentially enhance the augmentation process. Future work should investigate how these different versions of GAN and GPT influence performance metrics and learning outcomes, thereby improving the framework’s robustness, scalability, and adaptability across diverse ITS datasets. Additionally, we highlight that other open-source LLMs, such as Llama, Grok, and others, offer valuable opportunities for testing, whether through interface-based implementations or API versions. Secondly, the tensor-based data imputation and augmentation methods, although effective in this study, might not capture all nuances of learner behavior and performance. Tensor factorization models assume that learners at similar knowledge levels exhibit comparable performance patterns, which may not always hold true, especially in cases of highly individualized learning processes. Given the remarkable success of generative deep learning models, such as Generative Adversarial Imputation Networks (GAIN) \cite{yoon2018gain,zhang2024generative} and Autoencoders (AE) \cite{bank2023autoencoders} for data imputation through reconstruction mechanism, future work should explore their potential in handling data sparsity and capturing more complex patterns underlying learning performance data in ITSs. Thirdly, our study prioritizes the application of methodology over model optimization, examining how GenAI can enhance data augmentation. Although emerging advanced models, such as the GPT o-1, present promising avenues for further testing \cite{latif2024systematic}, limitations in file processing and project costs mean that a more extensive analysis is beyond the current scope and deferred to future work. Additionally, the broad scope of our systematic framework presents certain limitations, as it extends beyond a focused investigation of specific aspects such as multidimensional interactions, latent feature exploration in tensor factorization, and the verification of augmented data for real-world learner modeling. These areas may not have been explored in sufficient depth, limiting the granularity of insights. Lastly, while
this framework focuses on augmenting learning performance
data, it does not directly provide strategies for integrating the
proposed framework with comprehensive learner modeling in
ITSs to gain a more nuanced understanding of learner behavior
and enhance adaptive instruction. Future research should aim to address these gaps, enabling more effective personalization and support for learners within ITSs.

\section{Conclusions}

The present study presented a novel framework for augmenting learning performance data in ITSs by integrating multidimensional learner modeling (specifically tensor factorization) with GenAI models. The presented empirical evaluation suggests that this framework effectively densifies sparse learning performance data. We specifically addressed the challenge of learning performance data sparsity issues in ITSs using AutoTutor ARC as a case study, leveraging tensor factorization for data imputation and GenAI models, including Vanilla GAN and GPT-4o, for data augmentation. Our findings highlighted the superior predictive accuracy of tensor factorization over baseline methods, such as BKT, PFA and SPARFA-Lite, across four different AutoTutor ARC lessons across different difficulty levels. Furthermore, Vanilla GAN-based augmentations demonstrated desirable properties such as lower variance in parameter estimates. These improvements were largely reliable across different augmented sample sizes. Vanilla GAN-based augmentations exhibited desirable properties, including lower variance in parameter estimates, ensuring stable and reliable performance regardless of the total augmented sample size. In contrast, GPT-4o based augmentation demonstrated higher variability but occasionally showed closer alignment with the original data distribution, particularly in cases where bimodality was observed. These results underscore the complementary strengths of Vanilla GAN and GPT-4o: Vanilla GAN provides relatively reliable augmentation, while GPT-4o, though highly variable, offers the potential for replicating complex data structures and enhancing data diversity. The systematic augmentation framework effectively addresses data sparsity issues in ITSs, enhancing predictive capabilities and providing deeper insights into learners' progress. This framework has the potential to mitigate group and data selection biases that often occur during the collection of learning performance data in real-world experiments. The framework shows promise for enhancing ITSs through safe and cost-effective augmentation, enabling comprehensive data modeling by mitigating the impact of data sparsity, facilitating virtual evaluations of instructional designs, and paving the way for informed real-world applications to improve learners' outcomes in AI-driven education. 

\section*{Acknowledgments}

Our study was supported by the Institute of Education Sciences, U.S. Department of Education, through Grant R305A200413 and R305A190522 to The University of Memphis. The opinions expressed are those of the authors and do not represent views of the Institute or the U.S. Department of Education. We also extend our sincere gratitude to Prof. Philip I. Pavlik Jr. from the University of Memphis and Prof. Shaghayegh Sahebi from the University at Albany - SUNY for their expert guidance on the tensor factorization method. We also thank Prof. Mohammed Yeasin from the University of Memphis for his critical suggestions on the divergence measurement method used in this study. Additionally, we express our heartfelt thanks to Prof. Kenneth R. Koedinger of Carnegie Mellon University for his invaluable insights on knowledge tracing and simulations, which significantly enriched this study. 

\bibliographystyle{IEEEtran}
\bibliography{references.bib}

\section*{Biography Section}
\vspace*{-40pt}
\begin{IEEEbiography}[{\includegraphics[width=1in,height=1.25in,clip,keepaspectratio]{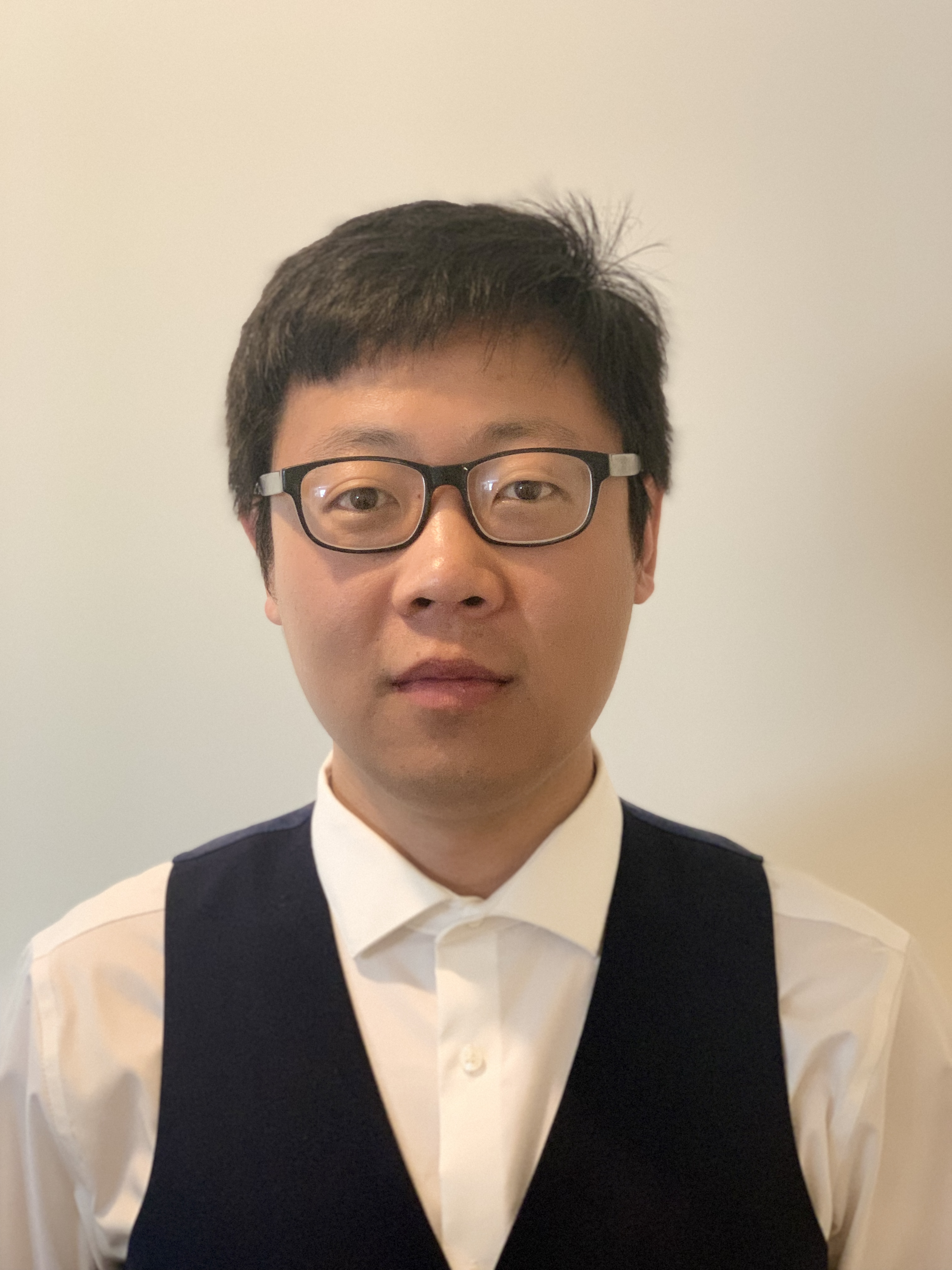}}]{Liang Zhang} is a PhD candidate in Computer Engineering at the Department of Electrical and Computer Engineering, University of Memphis. He currently serves as a research assistant at the Institute for Intelligent Systems (IIS) at University of Memphis and a visiting scholar at Educational Testing Service (ETS). He was an intern researcher at the Human-Computer Interaction Institute at Carnegie Mellon University from May to August 2023, and at the machine learning group at NEC Laboratories America from May to August 2024. His research focuses on human-computer interaction, including learning analytics, educational data mining, and GenAI for data imputation and augmentation in AI-education. He received the Best Full Paper Award at HCII’24. 
\end{IEEEbiography}
\vspace{-30pt}
\begin{IEEEbiography}[{\includegraphics[width=1in,height=1.25in,clip,keepaspectratio]{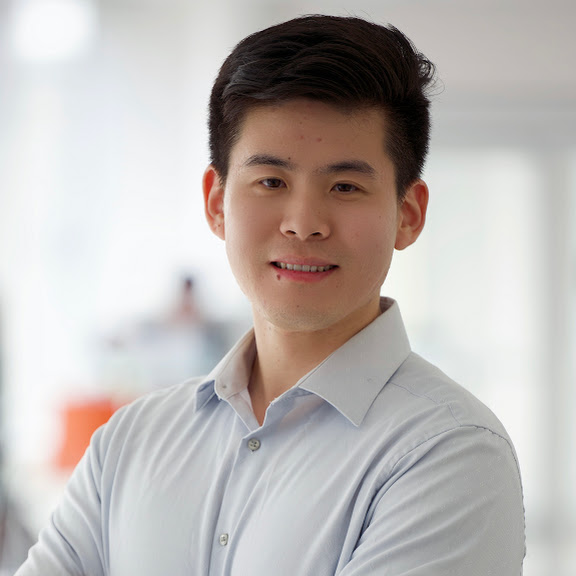}}]{Jionghao Lin} is an Assistant Professor in the Faculty of Education at the University of Hong Kong. Previously, he was a Postdoctoral Researcher at the Human-Computer Interaction Institute at Carnegie Mellon University, Pittsburgh, PA, USA, from 2023 to 2025. He received his Ph.D. in Computer Science from Monash University, Clayton, VIC, Australia, in 2023. His research interests include learning science, natural language processing, data mining, and applications of generative artificial intelligence in education. His work has been published in international journals and conferences and recognized with  awards, including the Best Paper Award at HCII'24, EDM'24, and ICMI'19, and the Best Demo Award at AIED'23.
\end{IEEEbiography}
\vspace{-30pt}
\begin{IEEEbiography}[{\includegraphics[width=1in,height=1.25in,clip,keepaspectratio]{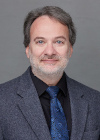}}]{John Sabatini} is a Distinguished Research Professor in the Department of Psychology and the Institute for Intelligent Systems at the University of Memphis. He was formerly a Principal Research Scientist at Educational Testing Service, specializing in reading literacy, assessment, cognitive psychology, and educational technology, with a focus on adults and adolescents.  
\end{IEEEbiography}
\vspace{-30pt}
\begin{IEEEbiography}[{\includegraphics[width=1in,height=1.25in,clip,keepaspectratio]{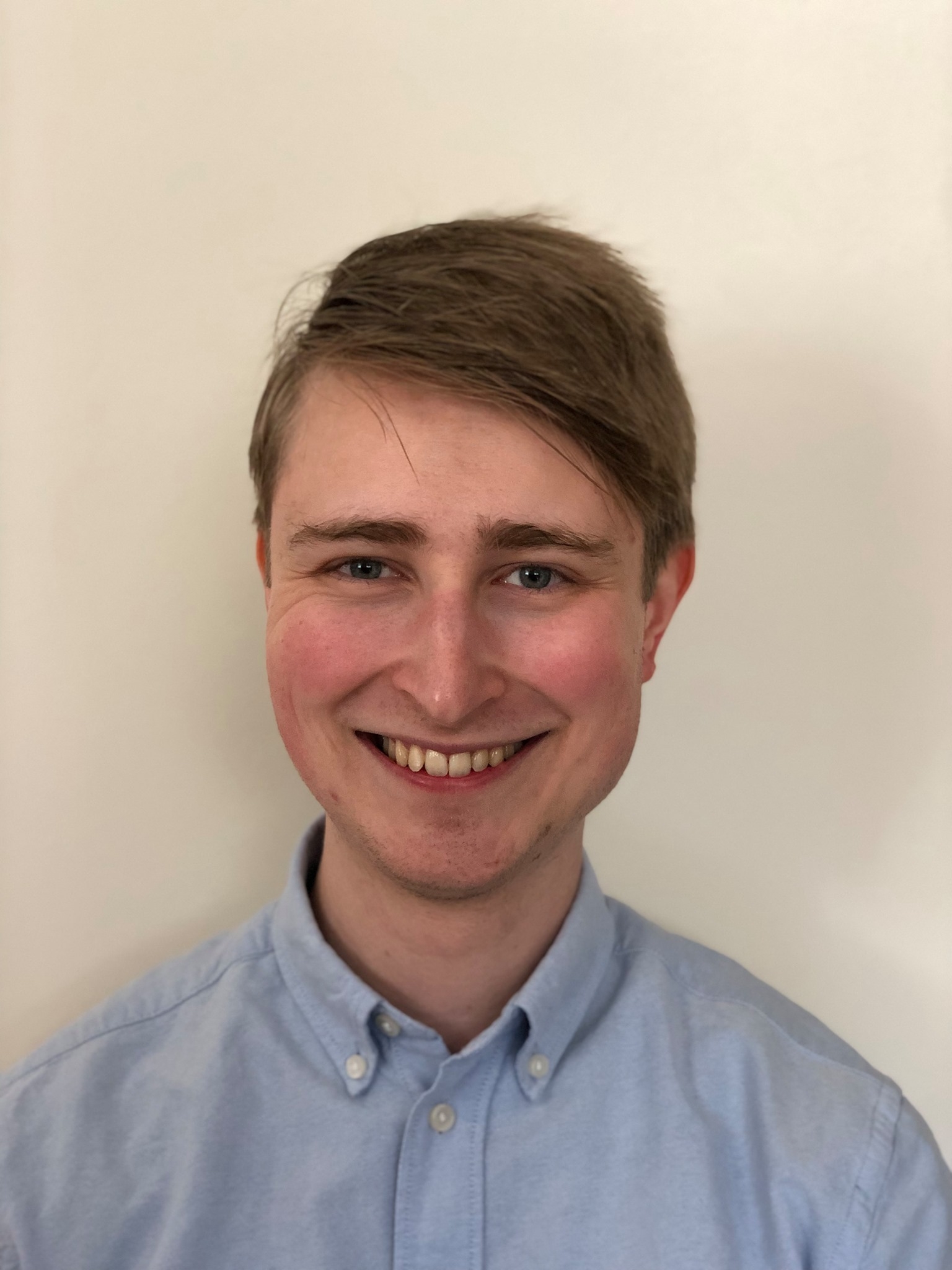}}]{Conrad Borchers} is a PhD student at the Human-Computer Interaction Institute (HCII) at Carnegie Mellon University’s School of Computer Science. He is broadly interested in studying the effectiveness of educational technologies and pathways through data science methods.  
\end{IEEEbiography}

\vspace{-30pt}
\begin{IEEEbiography}[{\includegraphics[width=1in,height=1.25in,clip,keepaspectratio]{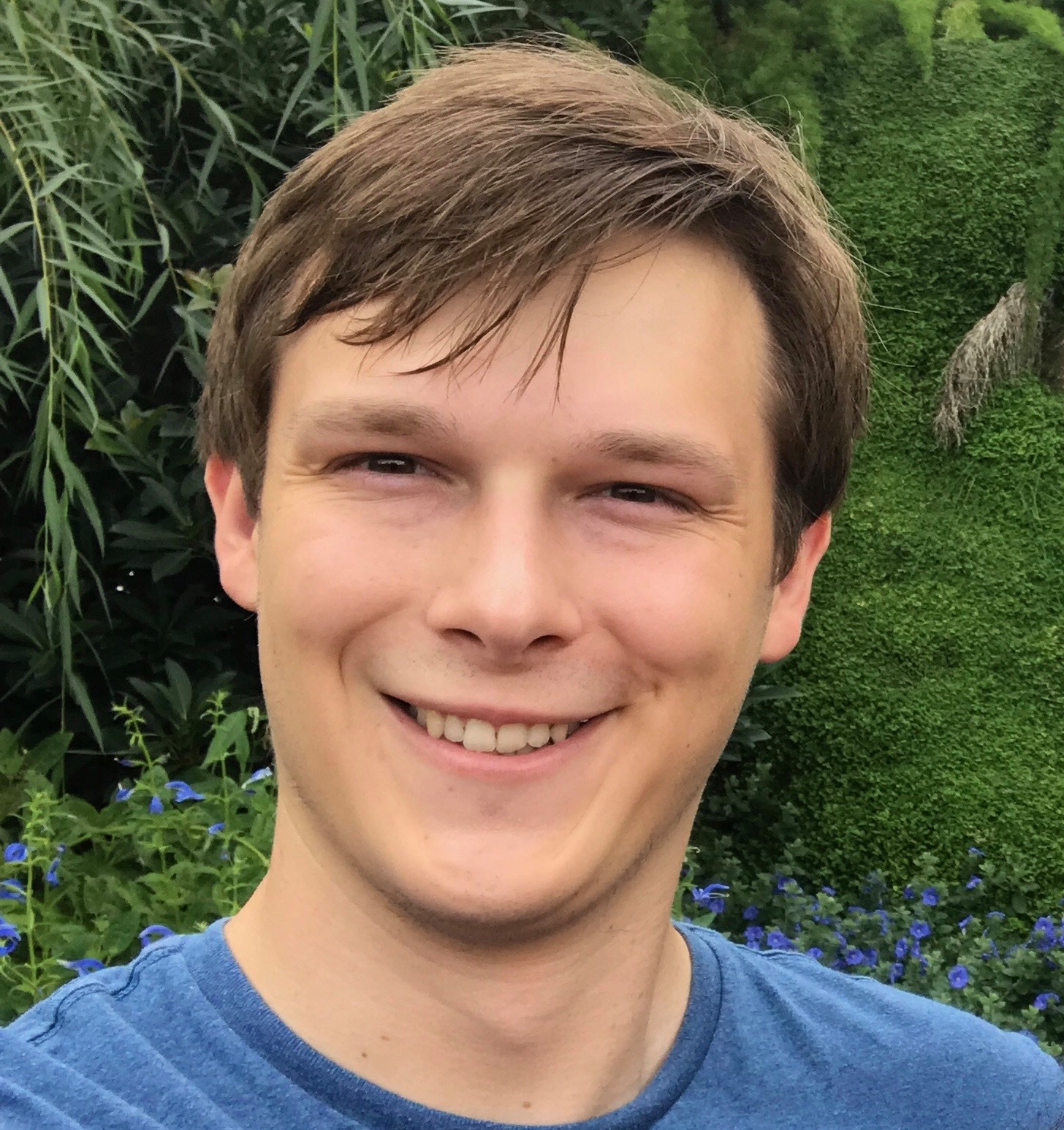}}]{Daniel Weitekamp} is a recent PhD graduate from Carnegie Mellon’s Human-Computer Interaction Institute. Daniel studies simulations of human learning that can learn directly from educational technology or interactively from a human instructor. His research has been applied to building computational models of learning, and for interactive task learning applications where non-programmers build whole programs by interactively teaching an AI. Daniel’s recent dissertation work shows that untrained users can build 100\% accurate intelligent tutoring systems in about 20 minutes by interactively teaching his AI, called AI2T.
\end{IEEEbiography}

\vspace{-30pt}
\begin{IEEEbiography}[{\includegraphics[width=1in,height=1.25in,clip,keepaspectratio]{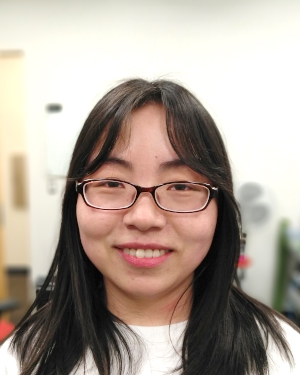}}]{Meng Cao} is a postdoctoral researcher in the Human-Computer Interaction Institute at Carnegie Mellon University, Pittsburgh, PA, USA. She received her Ph.D. in Experimental Psychology from the Department of Psychology at the University of Memphis. Her research interests include computational modeling, optimal learning, adaptive learning, and category learning.

\end{IEEEbiography}

\vspace{-30pt}
\begin{IEEEbiography}[{\includegraphics[width=1in,height=1.25in,clip,keepaspectratio]{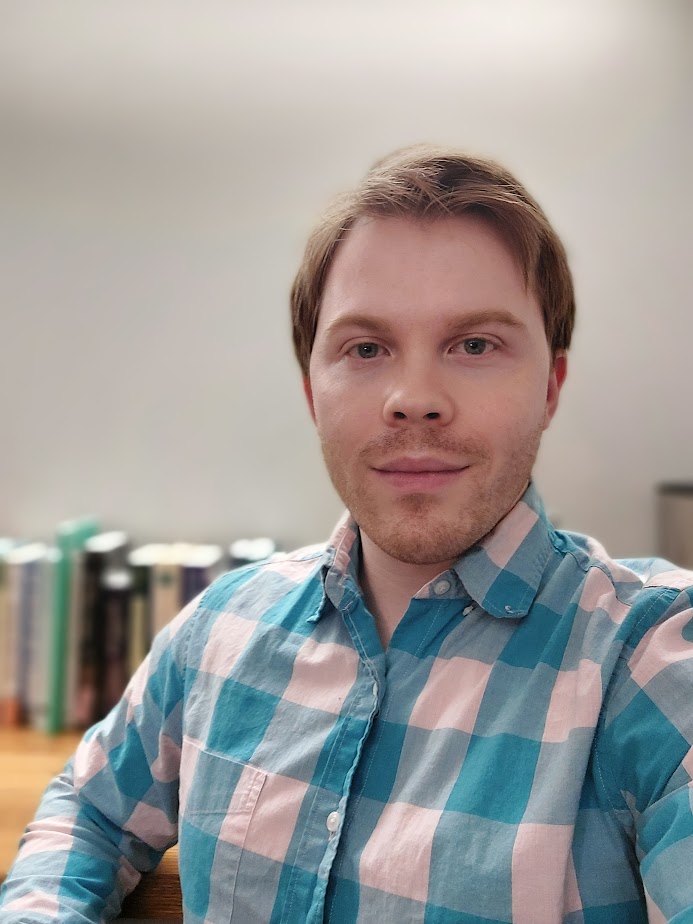}}]{John Hollander} is an Assistant Professor in the Department of Psychology and Counseling at Arkansas State University. His research interests include psycholinguistics, computational semantics, and the science of reading.

\end{IEEEbiography}
\vspace{-30pt}
\begin{IEEEbiography}[{\includegraphics[width=1in,height=1.25in,clip,keepaspectratio]{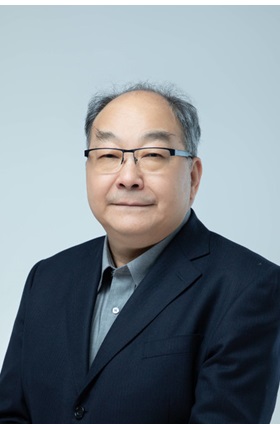}}]{Xiangen Hu} is Chair Professor of Learning Sciences and Technologies at Hong Kong Polytechnic University. He received his Doctor of Philosophy degree in Cognitive Psychology from the University of California, Irvine in 1991 and 1993 respectively. He took up an Assistant Professorship at The University of Memphis in 1993 and was promoted to Associate Professor in 2000 and Professor in the Department of Psychology in 2009. Meanwhile, he has been appointed Prof and Dean of the School of Psychology in the Central China Normal University since 2016 on a visiting basis where he has established good connections with the Chinese mainland. He joined PolyU in Dec 2023 as Director of the Institute for Higher Education Research and Development. His research areas include Mathematical Psychology, Research Design and Statistics, and Cognitive Psychology. 
\end{IEEEbiography}

\vspace{-30pt}
\begin{IEEEbiography}[{\includegraphics[width=1in,height=1.25in,clip,keepaspectratio]{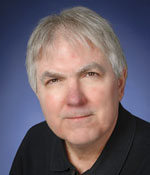}}]{Arthur C. Graesser} is a Distinguished University Professor in the Department of Psychology and the Institute of Intelligent Systems at the University of Memphis and is an Honorary Research Fellow in the Department of Education at the University of Oxford. He received his Ph.D. in psychology from the University of California at San Diego. Art's primary research interests are in cognitive science, discourse processing, and the learning sciences. More specific interests include knowledge representation, question asking and answering, tutoring, text comprehension, inference generation, conversation, reading, problem solving, memory, emotions, computational linguistics, artificial intelligence, human-computer interaction, and learning technologies with animated conversational agents. 
\end{IEEEbiography}

\flushbottom

\vfill

\end{document}